\documentclass[runningheads]{llncs}
\usepackage[T1]{fontenc}
\usepackage{graphicx}
\usepackage{booktabs}
\usepackage[misc]{ifsym}

\usepackage[sort&compress]{natbib}

\usepackage{amsmath,amsfonts,bm}

\def\1{\bm{1}}

\DeclareMathAlphabet{\mathsfit}{\encodingdefault}{\sfdefault}{m}{sl}
\SetMathAlphabet{\mathsfit}{bold}{\encodingdefault}{\sfdefault}{bx}{n}

\def\gG{{\mathcal{G}}}

\def\gO{{\mathcal{O}}}

\newcommand{\E}{\mathbb{E}}

\newcommand{\Var}{\mathrm{Var}}

\DeclareMathOperator*{\argmax}{arg\,max}
\DeclareMathOperator*{\argmin}{arg\,min}

\DeclareMathOperator{\Tr}{Tr}

\usepackage[utf8]{inputenc} 
\usepackage[T1]{fontenc}    
\usepackage{hyperref}       
\usepackage{url}            
\usepackage{booktabs}       
\usepackage{amsfonts}       
\usepackage{nicefrac}       
\usepackage{microtype}      
\usepackage{xcolor}         
\usepackage{amsmath}
\usepackage{amssymb}
\usepackage{mathtools}
\usepackage{amsthm}
\usepackage{comment}
\usepackage{algorithm}
\usepackage{algorithmic}
\usepackage{hyperref}
\theoremstyle{plain}
\usepackage{makecell}
\usepackage{subcaption} 
\newtheorem*{theorem*}{Theorem}
\theoremstyle{definition}

\theoremstyle{remark}
\usepackage{booktabs}
\usepackage{subcaption}
\usepackage{wrapfig}
\usepackage{multirow}
\newcommand{\zero}{\boldsymbol{0}}
\newcommand{\mean}{\boldsymbol{\mu}}
\newcommand{\cov}{\boldsymbol{\Sigma}}
\newcommand{\ecov}{\boldsymbol{S}}
\newcommand{\W}{\boldsymbol{W}}
\newcommand{\w}{\boldsymbol{w}}
\newcommand{\s}{\boldsymbol{s}}

\newcommand{\I}{\boldsymbol{I}}
\newcommand{\x}{\boldsymbol{x}}
\newcommand{\rX}{\mathbf{X}}

\newcommand{\Prob}{\mathbb{P}}
\newcommand{\M}{\mathcal{M}}
\newcommand{\N}{\mathcal{N}}

\newcommand{\nX}{\textnormal{X}}
\newcommand{\nY}{\textnormal{Y}}
\newcommand{\dX}{\boldsymbol{X}}
\newcommand{\dE}{\boldsymbol{E}}
\newcommand{\dEz}{\boldsymbol{E}_{\Z}}
\newcommand{\Z}{\mathbb{Z}}
\newcommand{\dZ}{\boldsymbol{Z}}
\newcommand{\dU}{\boldsymbol{U}}
\newcommand{\dN}{\N_{\Z}}
\newcommand{\dNt}{\N_{t\Z}}
\newcommand{\prcs}{\boldsymbol{\Theta}}

\usepackage{mwe}

\begin{document}

\title{Graph Structure Learning with Privacy Guarantees for Open Graph Data}
\titlerunning{Graph Structure Learning with Privacy Guarantees}
\author{Muhao Guo\inst{1} \and
Jiaqi Wu\inst{1} \and
Yizheng Liao\inst{1} \and
Wenke Lee\inst{2} \and
Shengzhe Chen\inst{1} \and
Yang Weng\inst{1}
}
\institute{
\inst{} Arizona State University, Tempe AZ 85281, USA \email{\{mguo26, jiaqiwu1, yzliao, schen415, Yang.Weng\}@asu.edu}
\and
\inst{} Georgia Institute of Technology, Atlanta, GA 30332, USA \email{wenke@cc.gatech.edu}
}



\maketitle            
\begin{abstract}
Publishing open graph data while preserving individual privacy remains challenging when data publishers and data users are distinct entities. Although differential privacy (DP) provides rigorous guarantees, most existing approaches enforce privacy during model training rather than at the data publishing stage. This limits the applicability to open-data scenarios. We propose a privacy-preserving graph structure learning framework that integrates Gaussian Differential Privacy (GDP) directly into the data release process. Our mechanism injects structured Gaussian noise into raw data prior to publication and provides formal $\mu$-GDP guarantees, leading to tight $(\varepsilon, \delta)$-differential privacy bounds. Despite the distortion introduced by privatization, we prove that the original sparse inverse covariance structure can be recovered through an unbiased penalized likelihood formulation. We further extend the framework to discrete data using discrete Gaussian noise while preserving privacy guarantees. Extensive experiments on synthetic and real-world datasets demonstrate strong privacy-utility trade-offs, maintaining high graph recovery accuracy under rigorous privacy budgets. Our results establish a formal connection between differential privacy theory and privacy-preserving data publishing for graphical models.
\end{abstract}


\section{Introduction}
\label{sec:intro}
Large-scale open data is critical to machine learning research, driving advancements across various domains \cite{auer2007dbpedia}. For example, ImageNet has been instrumental in shaping modern computer vision models \cite{deng2009imagenet}, while large language models such as Llama \cite{touvron2023llama} rely heavily on publicly available datasets. 
Open data broadens participation in AI research, enabling individuals and organizations to train and evaluate models without the constraints of proprietary data collection \cite{wu2019distributed}. Open data are structured records of user activities, including social network interactions \cite{volden2018legislative}, residential power meter data~\cite{liao2018urban,guo2023graph}, healthcare records \cite{cui2020coaid}, e-commerce transactions \cite{qiu2015predicting}, and online service histories \cite{bennett2007netflix}. 

In particular, data from the utility industry, such as residential power usage or grid data, offers significant value for research and development in smart grids, energy optimization, and sustainability efforts. By making such data publicly available, researchers can improve energy efficiency, reduce consumption, and optimize grid management. However, the accessibility of such sensitive data has been increasingly limited due to growing regulatory concerns and privacy challenges. For example, regulations like the GDPR impose stringent constraints on data publication, prohibiting the release of raw datasets without sufficient privacy safeguards \cite{voigt2017eu}.
This presents a significant challenge: how can we continue to leverage open data for research in fields like utility systems, while ensuring privacy protections for individuals?

One solution for this challenge is PPDP \cite{fung2010privacy}, which aims to release data in a way that preserves its utility while protecting individual privacy. Traditional anonymization techniques, such as k-Anonymity \cite{sweeney2002k}, l-Diversity \cite{machanavajjhala2007diversity}, and t-Closeness \cite{li2006t}, attempt to remove personally identifiable information before data publication. However, these methods have been shown to be vulnerable to attacks that leverage auxiliary knowledge \cite{domingo2015t}. In contrast, DP \cite{dwork2006calibrating} provides a mathematically rigorous privacy guarantee by ensuring that the presence or absence of any single individual in a dataset does not significantly change the outcome of an analysis, regardless of an attacker’s background knowledge \cite{mohammed2011differentially}. This makes DP a more robust approach to privacy preservation. However, existing DP methods are primarily designed for machine learning models that operate on private data internally, rather than for open data publishing scenarios where data must be shared externally while maintaining privacy. 
These challenges are further illustrated in Table \ref{tab:approach_comparison}, which provides a comparison of our proposed framework with existing privacy-preserving approaches, including traditional anonymization, DP, and PPDP.

Specifically, DP techniques primarily operate by adding noise during model training, ensuring that individual data points do not significantly influence learned parameters. For example, differentially private stochastic gradient descent (DP-SGD) \cite{abadi2016deep} injects noise into gradient updates, while matrix perturbation methods \cite{wang2018differentially} add noise to intermediate computations during training. Although these approaches ensure privacy in centralized machine learning models, they are not applicable to PPDP because the entity responsible for adding noise, i.e., the data publisher, is different from the data users who train models. This disconnect means that traditional DP mechanisms cannot be applied post-publication to protect privacy while maintaining analytical utility. Consequently, achieving DP under PPDP constraints, where raw data is shared in a privacy-preserving manner without compromising analytical validity, remains an open challenge.

Some approaches \cite{mckenna2022aim} add DP to data and thus perform DP-based privacy-preserving data release in general. However, the trade-off between privacy and performance persists. To address this dilemma, we concentrate on this subclass of problems, where the data publishers add noise to the raw data. In this setting, the data users, acting as a distinct party, can still achieve unbiased estimations from the noisy data with theoretical guarantees. 
A significant application of this approach is in graphical model structure learning, such as in the utility industry, where identifying relationships between different components of a power grid or energy consumption patterns can significantly improve operational efficiency. 
For example, in the utility industry, a sparse graph of power meter readings can help model and optimize energy consumption patterns.

Mathematically, learning the structure of a graphical model involves estimating a sparse inverse covariance matrix, where nonzero entries correspond to edges in the underlying graph. This problem is central to applications such as network inference, high-dimensional data analysis, and structured learning. A widely adopted approach for solving this problem is $\ell_1$-regularized log-determinant estimation, commonly known as graphical lasso \cite{friedman2008sparse}, which enforces sparsity in the estimated inverse covariance matrix. To preserve data privacy, prior works have attempted to introduce noise into either the covariance matrix or the learning process itself \cite{wang2018differentially}. However, these approaches fail to satisfy PPDP requirements, as they assume data users have direct access to raw data or summary statistics, which is often not feasible under strict privacy regulations. 

\begin{figure}[h!]
    \centering
    \includegraphics[width = \columnwidth]{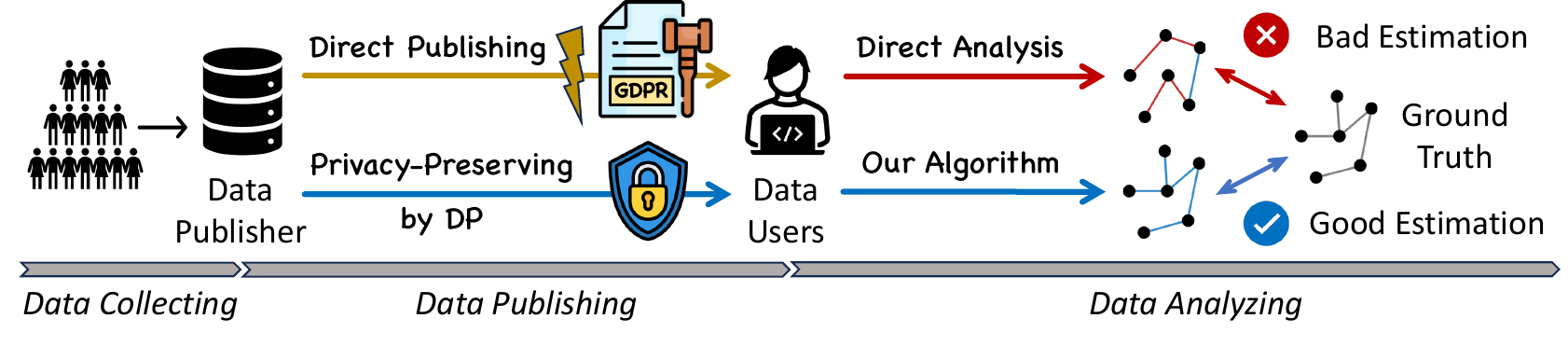}
    \caption{The data publisher can publish privacy-preserving data without sacrificing the data users' analysis performance.}
    \label{fig:Framework}   
\end{figure}

To bridge the gap between PPDP and DP, we propose a novel framework that ensures both privacy protection and accurate statistical estimation. As illustrated in Fig. \ref{fig:Framework}, our approach systematically integrates DP mechanisms at the data publishing stage, allowing data users to perform graphical model learning with mathematical guarantees. Unlike prior methods that introduce privacy at the model training stage, our approach directly modifies the raw data with structured noise, ensuring rigorous privacy guarantees while preserving essential statistical properties for unbiased learning. Specifically, we prove that our method satisfies $(\varepsilon, \delta)$-differential privacy while maintaining unbiased sparse inverse covariance estimation. To solve this privacy-preserving optimization problem, we develop modified versions of the coordinate descent \cite{friedman2008sparse} and alternating direction method of multipliers (ADMM) \cite{boyd2011distributed}. Furthermore, we extend our framework to handle discrete variables, a critical yet often overlooked aspect in privacy-preserving graphical model estimation, enabling applications in settings with categorical data, classification tasks, and structured decision processes. Unlike traditional input perturbation methods that treat privacy noise as an external preprocessing step, our framework provides an exact Gaussian Differential Privacy (GDP) characterization of publishing-stage perturbation and proves that sparse precision matrices can be recovered in an unbiased manner via post-processing invariance. This establishes a formal link between GDP theory and precision-matrix estimation under a distinct publisher–user model, which has not been characterized in prior graphical lasso literature.

Our contributions are threefold. 
First, we derive exact Gaussian Differential Privacy characterizations for covariance-based graph structure learning, providing closed-form $\mu$ expressions under both continuous and discrete Gaussian mechanisms. 
Second, we establish tight global sensitivity bounds and analyze diagonal and off-diagonal components to obtain worst-case GDP guarantees. 
Third, we connect covariance publishing with unbiased sparse inverse covariance estimation through post-processing invariance, enabling high-utility graph recovery under strict privacy constraints.
To the best of our knowledge, this is the first work providing exact GDP characterizations for covariance-based graph structure learning under both continuous and discrete Gaussian mechanisms.

\begin{table}[ht]
\centering
\caption{Comparison of Our Proposed Framework with Existing Privacy-Preserving Approaches}
\label{tab:approach_comparison}
\begin{tabular}{lcccc}
\toprule
\textbf{Approach} & \textbf{\begin{tabular}[c]{@{}c@{}}Privacy \\ Guarantee\end{tabular}} & \textbf{\begin{tabular}[c]{@{}c@{}}High Utility \\ (Open Data)\end{tabular}} & \textbf{\begin{tabular}[c]{@{}c@{}}Distinct \\ Pub/User\end{tabular}} & \textbf{\begin{tabular}[c]{@{}c@{}}Unbiased \\ Recovery\end{tabular}} \\ \midrule
Traditional Anonymization    & $\times$    & $\checkmark$ & $\checkmark$ & $\times$    \\
Internal DP (Model Training) & $\checkmark$ & $\times$    & $\times$    & $\times$    \\
Existing PPDP Methods        & $\checkmark$ & $\times$    & $\checkmark$ & $\times$    \\
\textbf{Our Proposed Framework} & $\checkmark$ & $\checkmark$ & $\checkmark$ & $\checkmark$ \\ \bottomrule
\end{tabular}
\end{table}

We validate our approach across diverse datasets that require strong privacy safeguards, including movie recommendation systems \cite{agarwal2011modeling}, brain networks \cite{yin2020gaussian}, cell signal analysis \cite{friedman2008sparse}, power systems \cite{liao2018urban}, social networks \cite{volden2018legislative}, disease spread modeling (chickenpox) \cite{rozemberczki2021chickenpox}, and soil microbiome interactions \cite{kurtz2019disentangling}.
Our results show that the proposed method can recover high-quality graph structures while preserving privacy, making it a viable solution for privacy-preserving data sharing in domains requiring sensitive information, such as the utility industry.
To further clarify the positioning of our framework relative to existing privacy-preserving approaches, Table~\ref{tab:approach_comparison} summarizes the key differences in privacy guarantees, utility preservation, and publishing assumptions.

\section{Related Work}

\noindent \textbf{Privacy protection in data publishing.}
Early approaches to privacy-preserving data publishing relied on anonymization techniques such as k-Anonymity \cite{sweeney2002k}, l-Diversity \cite{machanavajjhala2007diversity}, and t-Closeness \cite{li2006t}. These methods aim to remove or generalize identifying attributes before data release. However, it has been shown that anonymized datasets remain vulnerable to re-identification and linkage attacks when combined with auxiliary information \cite{narayanan2008robust}. Additional techniques, including data masking \cite{archana2018study} and data partitioning \cite{matatov2010privacy}, further obscure sensitive attributes but often introduce substantial information loss, particularly in high-dimensional settings. These limitations motivate privacy mechanisms with formal, adversary-independent guarantees.

\noindent \textbf{Differential privacy and its refinements.}
Differential privacy (DP) provides a mathematically rigorous notion of privacy that bounds the influence of any individual record on the released output \cite{dwork2006calibrating}. DP mechanisms typically add calibrated noise to query results or model parameters, enabling quantifiable privacy–utility trade-offs \cite{shokri2015privacy}. Owing to its strong adversarial guarantees, DP has been adopted in industrial systems \cite{erlingsson2014rappor} and governmental deployments \cite{haney2017utility}. Gaussian Differential Privacy (GDP) \cite{dong2019gaussian} further refines DP analysis by characterizing privacy guarantees through hypothesis testing trade-off functions, creating tight $(\varepsilon,\delta)$ bounds, and improved accounting under composition.

Despite these advances, most DP mechanisms are designed for centralized learning settings in which the data holder privately trains a model (e.g., via gradient perturbation or output perturbation) \cite{chaudhuri2011differentially, shokri2015privacy}. Such approaches do not directly address privacy-preserving \emph{data publishing}, where raw or privatized data must be released to external users under distinct trust boundaries.

\noindent \textbf{Differential privacy for graphical models.}
Sparse inverse covariance estimation and graphical lasso methods are fundamental tools for learning Gaussian graphical models in high-dimensional statistics \cite{dempster1972covariance, brillinger1996remarks, shrivastava2019glad}. Recent work has incorporated DP into graphical model estimation \cite{wang2018differentially}, typically by perturbing gradients, sufficient statistics, or model parameters during training. However, these methods generally assume centralized access to raw data and focus on private model training rather than publishing-stage protection. Existing DP graphical model approaches either privatize gradients during training or inject generic input noise without tight GDP characterization or unbiased precision recovery guarantees. Moreover, naive input perturbation can distort covariance structure and destroy sparsity patterns, leading to degraded or biased graph recovery \cite{li2023differentially}.

In contrast to prior approaches, we study privacy-preserving graph structure learning in a distinct publisher–user setting, where privatized data must be released while preserving downstream statistical validity. To our knowledge, existing work does not provide formal GDP guarantees together with an unbiased recovery formulation for sparse inverse covariance estimation under publishing-stage perturbation.

\section{Preliminaries}
\label{sec:prelim}
\subsection{Statistical Learning: Sparse Graph Estimation}
\label{sec:spare_related_works}
\label{sec:glasso}
Let $\gG = \{ V,E \}$ denote an undirected graph, where $V = \{1,2,\cdots,p \}$ represents the set of vertices and $E\subset V\times V$ represents the set of edges. Each vertex $i \in V$ corresponds to a random variable $\rX_i$. Such a structure, $\gG$, is recognized as a graphical model 
\cite{lauritzen1996graphical}. 
When each random variable follows a Gaussian distribution, 
we have a GGM. 
%
In a GGM, any two disconnected nodes, $\rX_i$ and $\rX_j$, are conditionally independent, meaning $\cov^{-1}_{ij} = 0$ for a multivariate Gaussian distribution $\rX \sim \mathcal{N}(\mean, \cov)$. 
Thus, sparse graph estimation is equivalent to finding the inverse covariance matrix, also known as the precision matrix, $\prcs = \cov^{-1}$. The most famous approach to estimating the precision matrix $\prcs$ is the graphical lasso~\cite{banerjee2008model}, which optimizes the penalized log-likelihood function:
\begin{align}
\label{graphical_lasso} 
\widehat{\prcs} \in \argmax\limits_{\Theta \succeq 0}  J(\Theta; \ecov),
\end{align} 
where $J(\Theta; \ecov) = \log\det (\Theta) - \Tr(\ecov\Theta) - \lambda \|\Theta\|_1$. Here, $\Theta \succeq 0$ denotes that $\Theta$ is positive semi-definite, $\log \det (\cdot)$ is the logarithm of the determinant, $\Tr(\cdot)$ is the trace operator, $\ecov$ is the empirical covariance matrix, $\lambda$ is the regularization parameter, and $\|\cdot\|_1$ denotes the $L_1$ norm 
for sparsity. Common methods to solve (\ref{graphical_lasso}) include coordinate descent~\cite{friedman2008sparse} and ADMM~\cite{boyd2011distributed}.

\subsection{Privacy-Preserving by Gaussian Differential Privacy}
\label{sec:def}
In this work, we employ DP \cite{dwork2006calibrating} defined as follows. 
%
%
\begin{definition}[\textbf{$(\varepsilon, \delta)$-Differential Privacy}]
A randomized algorithm $\mathcal{M}$ is said to satisfy $(\varepsilon, \delta)$-differential privacy if for any two neighboring datasets $X$ and $X'$ differing in exactly one element, and for all measurable sets $S \subseteq \mathrm{Range}(\mathcal{M})$,
\[
\Pr[\mathcal{M}(X) \in S] 
\le 
e^{\varepsilon} \Pr[\mathcal{M}(X') \in S] + \delta,
\]
where $\varepsilon \ge 0$ and $0 \le \delta \le 1$.
\end{definition}

Recent research indicates that Gaussian differential privacy (GDP) offers improved privacy-utility trade-offs and more natural privacy guarantees. Therefore, we introduce the following definition: 
\begin{definition}[\textbf{Trade-off function $T$}]
    For any two probability distributions $\mathbf{P}$ and $\mathbf{Q}$ in the same space, the difficulty of distinguishing $\mathbf{P}$ from $\mathbf{Q}$ by hypothesis testing is quantified by the trade-off function 
    $T (\mathbf{P}, \mathbf{Q}): [0, 1] \rightarrow [0, 1]$. It takes as input the level of type I error, $0<\alpha<1$, and outputs the corresponding type II error $\beta = T (\mathbf{P}, \mathbf{Q}) (\alpha)$.
\end{definition}
The trade-off function is a complete characterization of the fundamental difficulty in distinguishing two distributions through hypothesis testing~\cite{dong2019gaussian}. The greater this function, the harder it is to distinguish between the distributions.

\begin{definition}[Gaussian differential privacy]
A randomized algorithm $\mathcal{M}$ is said to satisfy $\mu$-Gaussian differential privacy ($\mu$-GDP) if for all neighboring datasets $X$ and $X'$ differing in exactly one element and for all $\alpha \in [0,1]$,
\[
T(\mathcal{M}(X), \mathcal{M}(X'))(\alpha) \;\ge\; G_{\mu}(\alpha),
\]
where $G_{\mu}(\alpha) := T(\mathcal{N}(0,1), \mathcal{N}(\mu,1))(\alpha)$ for $\mu \ge 0$.
\end{definition}

GDP is pivotal in this study for evaluating privacy preservation, leveraging its association with $(\varepsilon, \delta)$-DP. In the GDP framework, $T(\mathcal{M}(X),\mathcal{M}(X')) \geq G_\mu$ means to distinguish $\mathcal{M}(X)$ from $\mathcal{M}(X')$ is tougher than differentiating between normal distributions $\mathcal{N}(0,1)$ and $\mathcal{N}(\mu, 1)$ through hypothesis testing. If $\mathcal{M}$ is $\mu$-GDP, it is also $(\varepsilon, \delta)$-DP~\cite{dong2019gaussian}. Therefore, this GDP tool can bridge more intuitively to the classical DP framework.

\section{Threat Model and Privacy Assumptions}
\label{sec:threat_model}
To evaluate the robustness of our privacy-preserving estimation framework, we define a formal threat model that delineates trust boundaries, adversarial capabilities, and specific privacy targets relevant to open graph data publishing. This section serves to ground the mathematical guarantees of Gaussian Differential Privacy (GDP) within a practical security context, specifically addressing the challenges inherent when data publishers and users are distinct entities.

\paragraph{Threat Model.} We consider the central differential privacy setting. 
Two datasets $X,X'$ are defined as neighboring if they differ in exactly one entry.  We assume a bounded data domain $\|X\|_\infty \le B$, ensuring finite global sensitivity.  The adversary observes the privatized covariance matrix and attempts to distinguish between neighboring datasets.

\subsection{System Model and Roles}
We consider a decentralized data-sharing environment consisting of two primary roles: a trusted Data Publisher and an untrusted Data User who may also act as an Adversary. The Data Publisher is the sole entity with access to the raw, sensitive dataset $X \in \mathbb{R}^{n \times p}$ and is responsible for executing the Gaussian privatization mechanism to generate the noisy data $\tilde{X} = X + E$. Conversely, the Data User is an external third party who receives the privatized dataset $\tilde{X}$ and the associated noise strength $\sigma$. While the user's primary objective is to perform graphical model structure learning, they may also act as a passive adversary attempting to bypass privacy safeguards through statistical inference.

\subsection{Adversarial Capabilities and Knowledge}
We assume a passive, honest-but-curious adversary who adheres to data access protocols but attempts to extract sensitive individual information. This adversary possesses full observability of the released data, including the noisy data $\tilde{X}$ and the empirical covariance matrix $\tilde{S} = f(\tilde{X})$. Furthermore, we assume the adversary may possess significant auxiliary knowledge, such as information about all but one individual in the dataset, which is a standard assumption in the differential privacy framework to protect against linkage attacks with external datasets. Following the principle of methodological transparency, the adversary is assumed to know the privatization algorithm and the parameters $\sigma$ and $\mu$, though the specific noise realization $E$ remains hidden. This level of adversarial capability ensures that the privacy guarantees provided by the GDP framework are robust against sophisticated statistical reconstruction attempts.

\subsection{Privacy Objectives}
The primary privacy objectives of our framework are to defend against link disclosure and attribute inference attacks. In a link disclosure attack, the adversary attempts to determine whether a sensitive connection exists between two specific vertices $V_i$ and $V_j$ in the ground-truth graph $\mathcal{G}$. Our GDP mechanism ensures that the presence or absence of any single record in $X$ does not significantly alter the probability distribution of the published matrix $\tilde{S}$, making edge presence statistically indistinguishable from the noise injected during the privatization stage. Additionally, the framework prevents attribute inference, where an adversary attempts to reconstruct raw values from the published covariance matrix. By injecting structured noise $E$ that distorts the precision matrix while allowing for an unbiased global estimation $\hat{\Theta}$, we satisfy differential privacy requirements while maintaining high analytical utility for large-scale open data research and regulatory compliance.

\section{Methodology}
\label{sec:method}
We first propose our Gaussian privatization mechanism that preserves user privacy. We prove that such a method satisfies the differential privacy condition. Then, we propose our main result, the sparse graph estimation algorithm for privacy-preserving datasets. At last, we extend the proposed approach to handle the discrete datasets, which are common in machine learning tasks.

\subsection{Gaussian Data Privatization Mechanism}
For the multivariate Gaussian graphical model, we would like to recover its underlying structure via the graphical lasso algorithm introduced in section~\ref{sec:glasso}. Assume the raw data $\dX\in\mathbb{R}^{n\times p}$ is already centered by subtracting the mean. To recover the topology, one needs to compute statistics, in specific, the covariance matrix of $\dX$, which is $\ecov = f(\dX) = \frac{1}{n}\dX^{\top}\dX$. 
However, when the underlying data contains sensitive user data, it is dangerous to share the data with third parties directly. Therefore, we propose to have additive noises on the raw data for open data purposes: 
\begin{equation}
\label{eq:encrypt}
    \tilde{\dX} = \dX + \dE, 
\end{equation}
%
where $\tilde{\dX}$ denotes the privatized data. 
The noise term $\dE \in \mathbb{R}^{n\times p}$ has independent entries 
$\dE_{ij} \sim \mathcal{N}(0,\sigma^2)$. 
Under this privatization mechanism, the covariance matrix of $\tilde{\dX}$ becomes
\begin{equation}
    \label{eq:encryptcov}
    \tilde{\ecov} = f(\tilde{\dX}) = f(\dX) + \frac{1}{n}(\dE^{\top}\dX + \dX^{\top}\dE + \dE^{\top}\dE):=\mathcal{M}(\dX).
\end{equation}

Based on this covariance structure, we can establish the following theorem: a direct Gaussian privatization of the raw data satisfies $\mu$-Gaussian differential privacy (GDP), which is equivalent to $(\varepsilon,\delta)$-DP as defined in Section~\ref{sec:def}. This result yields a tight privacy bound and forms the basis for our unbiased estimator.


\begin{theorem}
\label{MDP}
Let $\mathcal{M}$ be the Gaussian privatization mechanism defined in
\eqref{eq:encryptcov}. Suppose the data matrix $X \in \mathbb{R}^{n \times p}$
satisfies the column-norm condition
\[
\frac{1}{n^2}\sum_{l=1}^n X_{lk}^2 \ge C_0 > 0
\quad \text{for all } k \in \{1,\dots,p\}.
\]
Then $\mathcal{M}$ satisfies $\mu$-Gaussian differential privacy ($\mu$-GDP) with
\[
\mu \;=\; \frac{\Delta_f}{\sigma \sqrt{2C_0}},
\]
where $\Delta_f \;=\; \sup_{X,X'} \| f(X) - f(X') \|$
is the global sensitivity of $f$ over all neighboring datasets $X$ and $X'$
that differ in at most one element. This sensitivity quantity is derived specifically for publishing-stage perturbation of the raw covariance data, rather than model-level noise injection, which distinguishes our setting from prior DP graphical lasso approaches.

Consequently, for any $\varepsilon \ge 0$, the mechanism $\mathcal{M}$ is
$(\varepsilon,\delta(\varepsilon))$-differentially private, where
\[
\delta(\varepsilon)
=
\Phi\!\left(-\frac{\varepsilon}{\mu} + \frac{\mu}{2}\right)
-
e^{\varepsilon}
\Phi\!\left(-\frac{\varepsilon}{\mu} - \frac{\mu}{2}\right),
\]
and $\Phi$ denotes the cumulative distribution function of the standard
normal distribution $\mathcal{N}(0,1)$.
\end{theorem}


Moreover, increasing the noise level $\sigma$ decreases $\mu=\Delta_f/(\sigma\sqrt{2C_0})$, strengthening privacy, but it also increases the distortion in the released covariance, inducing a formal privacy-utility trade-off.
The proof is in the Appendix~\ref{appendix:MDP}.

From the theorem, $\M$ is $\Delta_f(\sigma\sqrt{2C})^{-1}$-GDP. Such property ensures the utility of privacy preservation won't be under-estimated by any specific $(\varepsilon, \delta)-DP$ scheme. This is because GDP provides the tightest privacy bound~\cite{dong2019gaussian}. Using the GDP framework, we quantify the privacy-utility trade-off via the trade-off function $G_{\mu}$, which characterizes the optimal hypothesis testing error between $\mathcal{N}(0,1)$ and $\mathcal{N}(\mu,1)$. The value of sensitivity $\Delta_f$ can be determined by expertise when implementing in specific domains. As the strength of privacy utility is controlled by $\sigma$, a larger $\sigma$ makes distinguishing $\mathcal{N}(0,1)$ and $\mathcal{N}(\Delta_f(\sigma\sqrt{2C})^{-1},1)$ more difficult, thereby preserving more privacy.

\subsection{Sparse Graph Estimation with Privacy Preserving}

While the mechanism in (\ref{eq:encryptcov}) preserves the data privacy in a DP framework, the statistics of the privatized data $\tilde{\dX}$ no longer reflect the correct sparse graph connection.
Thus, there is no way to recover the topology merely according to $\tilde{\dX}$. 

\begin{theorem}
\label{tho:nontrivial}
Let $V=\{1,2,\dots,p\}$ be the set of graph nodes with $p>2$, and assume the graph is connected.
Let $\prcs = \cov^{-1}$ denote the precision matrix.
If $\sigma^2 < \|\prcs\|_2^{-1}$, where $\|\cdot\|_2$ denotes the spectral (operator) norm,
then for any $(i,j)$ such that $\prcs_{ij}=0$, we have
\[
\tilde{\prcs}_{ij}
=
\bigl(\cov + \sigma^2 \mathbf{I}\bigr)^{-1}_{ij}
\neq 0.
\]
\end{theorem}

The proof is in the Appendix~\ref{apendix:tho:nontrivial}.




Theorem~\ref{tho:nontrivial} shows that the original graph structure can be completely distorted even when only a small amount of Gaussian noise is added through the privatization mechanism. In particular, all zero entries of $\prcs$ become nonzero, meaning that sparsity no longer holds, and the graph structure induced by the perturbed precision matrix $\tilde{\prcs}$ becomes substantially denser. Consequently, one cannot trivially recover the original sparse graph structure from $\tilde{\prcs}$.
A similar conclusion was also suggested and validated in~\cite{10.5555/2834535}. This theorem confirms the effectiveness of using Gaussian noise as a privatization mechanism.

To address the issue above, in this section, we propose a new algorithm to estimate the original graph structure $\prcs$ with the privacy-preserving data $\tilde{\dX} = \dX + \dE$. As shown in Theorem~\ref{tho:nontrivial}, since each row of data now follows $\N(\zero,\cov + \sigma^2 \I)$, using the ordinary graphical lasso in (\ref{graphical_lasso}) yields an estimation of $\tilde{\prcs}$, whose graph structure is significantly distorted. However, with minimal additional information, we are able to recover the original graph structure $\prcs$ based on privatized data $\tilde{\dX}$. The following theorem guarantees the equivalent recovery of graph structure with privatized data and privatized empirical covariance matrix $\tilde{\ecov} = \frac{1}{n}\tilde{\dX}^{\top}\tilde{\dX}$. 


\begin{theorem}[\textbf{Sparse graph estimation with continuous privacy-preserving data}]
\label{tho:pgl}
Given $\tilde{\ecov},\sigma$, the penalized log-likelihood function
$\displaystyle J(\Theta;\tilde{\ecov}-\sigma^2\I)$ 
is an unbiased estimator of $J(\Theta; \ecov)$, in other word:
\begin{equation}
    J(\Theta; \ecov) =  \E\left[J(\Theta;\tilde{\ecov}-\sigma^2\I)\right].
\end{equation}
\end{theorem}
The proof is in the Appendix~\ref{apendix:tho:pgl}.

This theorem shows that $\displaystyle J(\Theta;\tilde{\ecov}-\sigma^2\I)$ is an unbiased approximation of the vanilla penalized log-likelihood $J(\Theta;\ecov)$. Therefore, maximizing this estimator simultaneously maximizes the vanilla graphical lasso in~(\ref{graphical_lasso}). The only additional information required is the noise level $\sigma$, which specifies the strength of the Gaussian perturbation used in the privatization mechanism. Consequently, we propose a privacy-preserving graphical lasso algorithm to recover the underlying graph structure:

\begin{equation}
\label{estimation}
\widehat{\prcs}\in \argmax\limits_{\Theta \succeq 0}J(\Theta;\tilde{\ecov}-\sigma^2\I).
\end{equation}

Corollary \ref{corollary: Uniqueness of prcs} ensures $\widehat{\prcs}$ is also a solution to the vanilla lasso problem (\ref{graphical_lasso}).
\begin{corollary}[\textbf{Uniqueness of $\widehat{\prcs}$}] According to Theorem~\ref{tho:pgl}, $\widehat{\prcs}$ maximizes  
$J(\Theta; \ecov),J(\Theta;\tilde{\ecov}-\sigma^2\I)$
simultaneously.
    Since both objectives are strictly concave with respect to $\Theta$ over the positive definite cone, they admit unique maximizers. Thus, $\widehat{\prcs}$ is also the unique maximum point of $J(\Theta; \ecov)$.
\label{corollary: Uniqueness of prcs}
\end{corollary}

To solve (\ref{estimation}), we can utilize coordinate descent~\cite{friedman2008sparse} in Algorithm~\ref{alg:cd} or ADMM~\cite{boyd2011distributed} in Algorithm~\ref{alg:admm}, depending on specific scenarios. Coordinate descent is an efficient optimization method known for its speed; however, it may not be suitable for all types of optimization problems. Alternatively, we can apply ADMM, which is more stable but requires more computations until convergence. We will validate (\ref{estimation})'s effectiveness in recovering graph structure, implemented to multiple domains in the numerical experiments section.

\begin{algorithm}
\caption{Sparse Graph Estimation with Privacy-Preserving Data by Coordinate Descent}
\label{alg:cd}
\leftskip 0mm
\fontsize{8pt}{8pt}\selectfont
\textbf{Input:} Privatized data $\tilde{\dX} = \dX + \dE$ by (\ref{eq:encrypt}), strength of noise $\sigma$. 

\textbf{Hyper-parameters:} Penalty $\lambda>0$ for $L_1-$ regularization.

\textbf{Initialization}. $\widehat{\prcs}\in \mathbb{R}^{p\times p}$.  Privatized covariance matrix $\tilde{\ecov} = \frac{1}{n}\tilde{\dX}^{\top}\tilde{\dX}$. 
Refined matrix $\widehat{\ecov} = \tilde{\ecov}-\sigma^2\I$.
$\W = \widehat{\ecov} + \lambda \I$. The diagonal of $\W$ remains unchanged in what follows. 

\begin{algorithmic}[1]
\FOR{$j = 1,2,\cdots,p,1,2,\cdots,p,\cdots$}
\STATE Partition $\W,\widehat{\ecov}$ into 4 parts: all but the $j$-th row and column, $j$-th row and column but the $j$-th element in diagonal, and the $j$-th element in diagonal:
\begin{equation}
    \W=\left[\begin{array}{cc}
        \W_{11} & \w_{12} \\
        \w_{12}^T & \w_{22}
    \end{array}\right], \ \widehat{\ecov}=\left[\begin{array}{cc}
        \ecov_{11} & \s_{12} \\
        \s_{12}^T & \s_{22}
    \end{array}\right].
\end{equation}
\STATE  Solve the lasso problem, where $\boldsymbol{b} = \W_{11}^{-1/2} \s_{12}$:
    $
    \hat{\boldsymbol{\beta}} = \argmin_{\boldsymbol{\beta}}  \frac{1}{2} \left\Vert \W_{11}^{1/2} \boldsymbol{\beta} - \boldsymbol{b} \right\Vert_2^2 + \lambda \left\Vert\boldsymbol{\beta}\right\Vert_1
    $
\STATE Update $\w_{12} = \W_{11}\hat{\boldsymbol{\beta}}$. Store $\boldsymbol{\beta}_{j} = \hat{\boldsymbol{\beta}}$. Check convergence.
\ENDFOR
\FOR{$j = 1,2,\cdots,p$}
\STATE Partition $\widehat{\prcs}$ as the same way with $\W,\widehat{\ecov}$ into 4 parts according to $j$-th element in diagonal.
\STATE The $j$-th element in diagonal $\widehat{\theta}_{22} = 1/\left( \w_{22} - \w_{12}^T\boldsymbol{\beta}_{j} \right)$.
\STATE The $j$-th raw $\widehat{\theta}_{12} = -\boldsymbol{\beta}_{j}\widehat{\theta}_{22}$. The $j$-th column is its transpose.
\ENDFOR
\end{algorithmic}
\textbf{Output:}  $\widehat{\prcs}$, the estimated graph structure.
\end{algorithm}


\begin{algorithm}[t]
\caption{ADMM for Graphical Lasso on Privatized Covariance}
\label{alg:admm}
\begin{algorithmic}[1]
\FOR{$k = 1,2,\cdots, K$}
\STATE
$
\prcs^{k+1} = \arg\min_{\Theta}
\left\{
-\log\det(\Theta) + \Tr(\widehat{\ecov}\Theta)
+ \frac{\rho}{2}\left\Vert \Theta - \dZ^{k} + \dU^{k} \right\Vert_{F}^2
\right\}
$
\STATE
$
\dZ^{k+1} = \arg\min_{\dZ}
\left\{
\lambda \Vert\dZ\Vert_1
+ \frac{\rho}{2}\left\Vert \prcs^{k+1} - \dZ + \dU^{k} \right\Vert_{F}^2
\right\}
$
\STATE
$\dU^{k+1} = \dU^{k} + \prcs^{k+1} - \dZ^{k+1}$.
\STATE Check convergence.
\ENDFOR
\end{algorithmic}
\textbf{Output:} Recovered graph structure $\widehat{\prcs} = \prcs^{K}$.
\end{algorithm}
Each iteration of Algorithm~2 requires solving a matrix inverse or eigen-decomposition for the $\Theta$-update, resulting in a computational complexity of $\mathcal{O}(p^3)$ per iteration.

\subsection{DP and Sparse Graph Estimation with Discrete Data}
In the previous sections, we addressed the challenges of sparse graph estimation with continuous and privacy-preserving data. However, in many practical applications, such as social networks, the collected data is in the discrete domain. Applying Gaussian noise to discrete variables can significantly hinder interpretability \cite{canonne2020discrete}. 
In this section, we discuss how to preserve data privacy for discrete Gaussian variables $\rX_k \in \mathcal{X}$ and analyze the utility of the proposed privatization algorithm obtained by adding discrete Gaussian noise.

\begin{definition}[\textbf{Discrete Gaussian}\cite{canonne2020discrete}]
    Discrete random variable $\rX_k$ following discrete Gaussian distribution is denoted as $\rX_k\sim\dN(\mu,\sigma^2)$, with probability at each $x\in\Z$:
    \begin{equation}
    \label{eq:dis_gaussian}
        \Prob\left[\rX_k = x\right] = \frac{\exp\left\{ - (x-\mu)/(2\sigma^2) \right\}}{\sum_{y\in\Z}\exp\left\{ - (y-\mu)/(2\sigma^2) \right\}}.
    \end{equation}
\end{definition}
(\ref{eq:dis_gaussian}) is a natural discretization of $\N(\mu,\sigma^2)$. 
Now, suppose our privatization mechanism in (\ref{eq:encrypt}) adds a discrete Gaussian noise to the centered raw data $\dX$, i.e. $\tilde{\dX} = \dX + \dE_{\mathbb{Z}}$, where components of $\dE_{\mathbb{Z}}$ are i.i.d. $\dN(0, \sigma^2)$. We have the following theorem to evaluate $\M$'s privacy utility in (\ref{eq:encryptcov}).

\begin{theorem}
\label{tho:DMDP}
     In the case of adding discrete Gaussian noise as a privatization mechanism, $\M$ is $\frac{\Delta_f + 2K}{\sqrt{2n}K\sigma}$-GDP.  It is also $(\varepsilon, \delta(\varepsilon))$-DP for all $\varepsilon\ge0$,
        $\delta(\varepsilon) = \Phi\left(-\frac{\varepsilon\sqrt{2n}K\sigma}{(\Delta_f + 2K)} + \frac{\Delta_f + 2K}{2\sqrt{2n}K\sigma} \right) - e^{\varepsilon}\Phi\left(-\frac{\varepsilon\sqrt{2n}K\sigma}{(\Delta_f + 2K)} - \frac{\Delta_f + 2K}{2\sqrt{2n}K\sigma} \right)$,
    where $\Phi$ is the CDF of $\N(0,1)$, $K = \frac{1}{n}\min_{\x_{li}\in\dX}|\x_{li}|$, and $\Delta_f$ is the sensitivity of two neighboring data matrices differing in exactly one element.
\end{theorem}
The proof is in the Appendix~\ref{apendix:tho:DMDP}.

When it comes to sparse graph estimation under discrete Gaussian privatization, we have the following theorem.
\begin{theorem}[\textbf{Sparse graph estimation with discrete privacy-preserving data}]
\label{tho:pgld} 
In the case where each entry of $\dX$ is perturbed with discrete Gaussian noise $\mathcal{N}(\mathbf{0}, \sigma^2)$, let $\tilde{\ecov}$ be the empirical covariance matrix, $\bar{\sigma}=\Var[\dN(0, \sigma^2)]$, then the penalized log-likelihood function $\displaystyle J(\Theta;\tilde{\ecov}-\bar{\sigma}\I)$ is an unbiased estimator of $J(\Theta; \ecov)$. 
\begin{equation}
\label{estimationdis}
    J(\Theta; \ecov) =  \E\left[J(\Theta;\tilde{\ecov}-\bar{\sigma}\I)\right].
\end{equation}
\end{theorem}


First, we have the proposition about the expectation and variance of discrete Gaussian noise:
\begin{proposition}[Expectation and variance of discrete Gaussian~\cite{canonne2020discrete} ]
\label{prop2}
For $\nX \sim\dN(\mu,\sigma^2)$, $\E[\nX] = \mu$. Let $\bar{\sigma}=\Var[\nX]$,
\begin{equation} 
\bar{\sigma} \le \sigma^2 \left(1- \frac{4\pi^2\sigma^2}{\exp\{4\pi^2\sigma^2\} - 1} \right) < \sigma^2.
\end{equation}
\end{proposition}

Then, the proof is similar to the proof of Theorem~\ref{tho:pgl}.
However, when it comes to calculate $\E[\dE_{li}\dE_{lj}]$, in the case $i=j$, $\E[\dE_{li}\dE_{lj}] = \bar{\sigma}$.

In practice, we can estimate $\bar{\sigma}$ by sampling sufficiently large volume from $\dN(0, \sigma^2)$. Therefore, to recover graph structure, we can still utilize Algorithms~\ref{alg:cd} and ADMM~\ref{alg:admm}, but instead substitute the refined matrix by $\displaystyle\widehat{\ecov} = \tilde{\ecov}-\bar{\sigma}\I$.

\section{Numerical Validation}
\label{sec:numexp}
We evaluate our approach on several synthetic datasets and eight real-world datasets, including six continuous and three discrete datasets. 
Considering the effect of random noise, we run each experiment $10$ times and record the corresponding mean and standard deviation.
All the models were implemented in Python 3.9. All experiments were conducted using a device equipped with an M2 processor featuring an 8-core CPU. Our code is available at \href{https://github.com/MuhaoGuo/Structure_Learning_Graphical_Lasso}{\texttt{Github}}.



\subsection{Evaluation Metrics}
The sparse graph estimation 
can be framed as a binary classification problem for each potential edge. 
So, we use Receiver Operating Characteristic (ROC) curves and the Area Under the Curve (AUC) score as evaluation metrics. In cases where a dataset lacks a ground truth structure, we adopt the outcome of the standard graphical lasso as the de facto ground truth 
to check if our model can achieve performance similar to a vanilla graphical lasso.


\subsection{Cross-validation for Selecting $\lambda$}
\label{Appendix: Parameter tunning}
Cross-validation is an effective strategy to select the parameter $\lambda$ \cite{friedman2008sparse}. This method involves dividing the data into $K$ folds and iteratively training the graphical lasso model on $K-1$ folds while the remaining fold serves for validation across a spectrum of $\lambda$ values.  The selection process utilizes the mean squared error (MSE) between the predicted and actual covariance matrices from training and validation sets. The objective is to identify an optimal $\lambda$ that ensures a balance between minimizing MSE and maintaining a graph structure with meaningful edges. Selecting the $\lambda$ that minimizes MSE without resulting in an overly sparse graph devoid of any edges is crucial. The optimal $\lambda$ is highlighted based on achieving a low MSE while preventing the graph from becoming too sparse.
\begin{figure}[h!]
    \centering
    \includegraphics[width = 0.8\columnwidth, trim=4 4 4 4,clip]{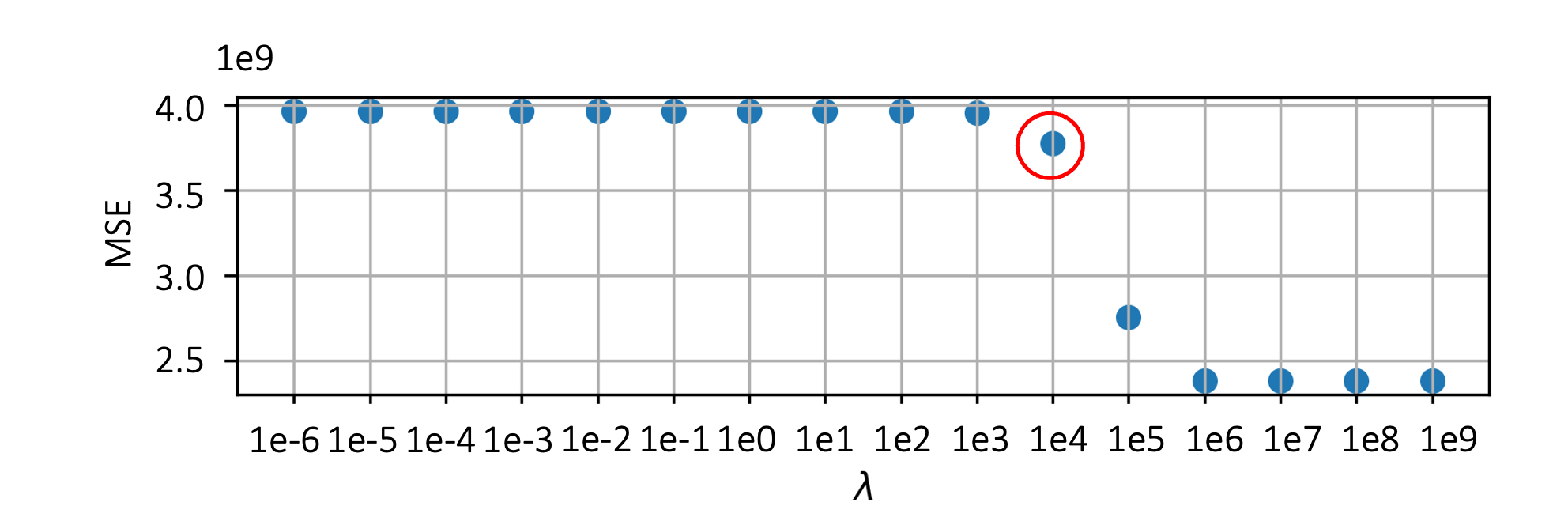}
    \caption{The cross-validation for selecting $\lambda$. The red circle represents the corresponding MSE of the selected $\lambda$.}
    \label{cross_validation}
\end{figure}

\subsection{Complexity Analysis}
\label{sec: Complexity analysis}
Let $n$ be the number of samples (rows in $X$) and $p$ be the number of variables (columns in $X$). The complexity includes two parts. The privatization complexity is $\mathcal{O}(np)$, as we add Gaussian noise to the data matrix $\mathbb{R}^{n\times p}$. The second part involves applying the reconstruction algorithm to recover the graph from the empirical covariance matrix $S \in \mathbb{R}^{p \times p}$.
The complexity of this part depends on the method used, such as Coordinate Descent or ADMM. So, let $k$ be the number of iterations required for convergence. The total complexity of the iterative part is $O(kp^3)$, not considering the benefits of sparsity. It is important to note that methods like Coordinate Descent or ADMM are necessary for similar reconstruction methods, so our approach does not incur additional computational costs.

\subsection{Privacy Protection}

We use signal-to-noise ratio (SNR) in decibel (dB) to measure the noise level applied to the data for obfuscation \cite{mivule2013utilizing}. Let $\Delta_f = \sup_{\dX,\dX'} \| f(\dX) - f(\dX') \|$
denote the global sensitivity of $f$ over neighboring datasets $\dX$ and $\dX'$ that differ in at most one element.
For covariance estimation under bounded data,
\[
\Delta_f = \frac{2 ( \max(\dX) - \min(\dX) )^2}{n}.
\]

Given noise standard deviation $\sigma$, the Gaussian privatization mechanism satisfies $\mu$-Gaussian differential privacy with
\[
\mu = \frac{\Delta_f}{\sigma \sqrt{2 C_0}},
\]
as established in Theorem~5.1. We then compute $(\varepsilon,\delta)$ numerically using the GDP-to-DP conversion
\[
\delta(\varepsilon)
=
\Phi\!\left(-\frac{\varepsilon}{\mu} + \frac{\mu}{2}\right)
-
e^{\varepsilon}
\Phi\!\left(-\frac{\varepsilon}{\mu} - \frac{\mu}{2}\right),
\]
for a desired $\delta$ level. Table~\ref{AUC} reports the resulting $\varepsilon$ values corresponding to each SNR level for each dataset.

\subsection{Synthetic and Large Dataset}
\label{sec: Synthetic and Large Dataset}


We first generate random networks of $50$ nodes.
Each node's values were randomly sampled from a multivariate Gaussian distribution. 
We generate the precision matrix following the work of \cite{yuan2007model} by: $(\prcs)_{ii} = 1, (\prcs)_{i,i-1} = (\prcs)_{i-1,i} = 0.5$, and zero otherwise.
We use the vanilla graphical lasso on the non-privatized data to establish a baseline graph structure, considered as the ground truth. Then, we add various 
levels of noise. 
We observe from Table~\ref{tab:synthetic} that our method is not only scalable but also robust to different noise levels. 
We next tested our method on a large dataset consisting of $1,000$ nodes and $10,000$ samples, with varying levels of sparsity. 
Sparsity is defined as the ratio of zero-valued elements to the total number of elements in the matrix. Similarly, we created a precision matrix $\prcs$ with size $1,000 \times 1,000$. We set $(\prcs)_{i,j} = (\prcs)_{j, i} = 0.1$, where $i\neq j$, and $(\prcs)_{i,i} = \sum((\prcs)_i)$. We calculated the covariance of  $\prcs$ by matrix inversion. Once we obtained the covariance matrix, we set the mean to $0$ and generated a dataset following the multivariate normal distribution. The performance of our method at different sparsity ratios is shown in Table \ref{table:large_toy_data_sparsity_ratios}. 
We found that performance improves with more sparse networks. 


\begin{table*}[h]
\centering
\begin{subtable}[t]{0.6\linewidth}
    \centering
    \small
    \caption{AUC of our approach with varying sample sizes on synthetic data.}
    \resizebox{\linewidth}{!}{
    \begin{tabular}{c|c|c|c}
        \toprule
        Sample Size & 50 & 500 & 5000 \\
        \midrule
        \midrule
        No noise & $1.0\pm0.0$ & $1.0\pm0.0$ & $1.0\pm0.0$ \\
        \midrule
        100 dB   & $0.999\pm2.3\text{e-5}$ & $0.999\pm1.8\text{e-3}$ & $0.998\pm1.8\text{e-3}$ \\
        80 dB    & $0.999\pm1.0\text{e-5}$ & $0.999\pm2.0\text{e-3}$ & $0.998\pm3.2\text{e-3}$ \\
        60 dB    & $0.998\pm5.2\text{e-3}$ & $0.999\pm1.9\text{e-3}$ & $0.997\pm1.2\text{e-2}$ \\
        40 dB    & $0.989\pm4.7\text{e-3}$ & $0.993\pm5.7\text{e-3}$ & $0.994\pm1.1\text{e-2}$ \\
        20 dB    & $0.884\pm2.6\text{e-2}$ & $0.934\pm2.1\text{e-2}$ & $0.949\pm1.4\text{e-2}$ \\
        \bottomrule
    \end{tabular}
    }
    \label{tab:synthetic}
\end{subtable}
\begin{subtable}[t]{0.8\linewidth}
    \centering
    \caption{AUC of the large dataset with 1,000 nodes and different sparsity ratios.}
    \resizebox{\linewidth}{!}{
    \begin{tabular}{c|c|c|c|c}
        \toprule
        Sparsity & 99\% & 98\% & 95\% & 90\% \\
        \midrule
        \midrule
        No noise & $0.999\pm 7.51\text{e-4}$ & $0.998\pm 2.56\text{e-4}$  & $0.835\pm 7.51\text{e-4}$ & $0.626\pm 7.51\text{e-4}$ \\
        \midrule
        100 dB & $0.999\pm 7.51\text{e-4}$ & $0.998\pm 2.56\text{e-4}$ & $0.835\pm 7.51\text{e-4}$ & $0.626\pm 7.51\text{e-4}$ \\
        80 dB & $0.999\pm 7.51\text{e-4}$ & $0.998\pm 2.56\text{e-4}$ & $0.835\pm 7.51\text{e-4}$ & $0.626\pm 7.51\text{e-4}$ \\
        60 dB & $0.999\pm 7.51\text{e-4}$ & $0.998\pm 2.56\text{e-4}$ & $0.835\pm 7.51\text{e-4}$ & $0.626\pm 7.50\text{e-4}$ \\
        40 dB & $0.999\pm 7.54\text{e-4}$ & $0.998\pm 2.56\text{e-4}$ & $0.835\pm 7.54\text{e-4}$ & $0.625\pm 7.49\text{e-4}$ \\
        20 dB & $0.999\pm 8.87\text{e-4}$ & $0.998\pm 3.20\text{e-4}$ & $0.830\pm 9.08\text{e-4}$ & $0.623\pm 8.54\text{e-4}$ \\
        \bottomrule
    \end{tabular}
    }
    \label{table:large_toy_data_sparsity_ratios}
\end{subtable}
\end{table*}




\subsection{Applications on Real-World Datasets}
\begin{figure}[t!]
    \vspace{-3mm}
    \includegraphics[width = 1.0\columnwidth]{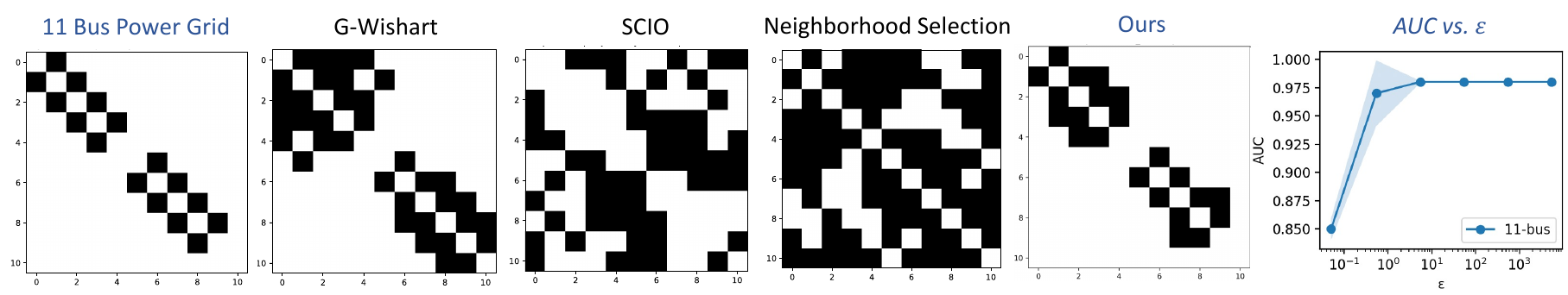}
    \caption{Estimated adjacency matrices for an 11-bus power grid at 20 dB noise. Left to right: ground truth, G-Wishart, SCIO, Neighborhood Selection, our approach, and its privacy–utility curve.}
    \label{baselines}
    \vspace{-3mm}
\end{figure}

\begin{figure*}[t]
    \centering
    \includegraphics[width = 0.97 \columnwidth, trim=4 4 4 4, clip]{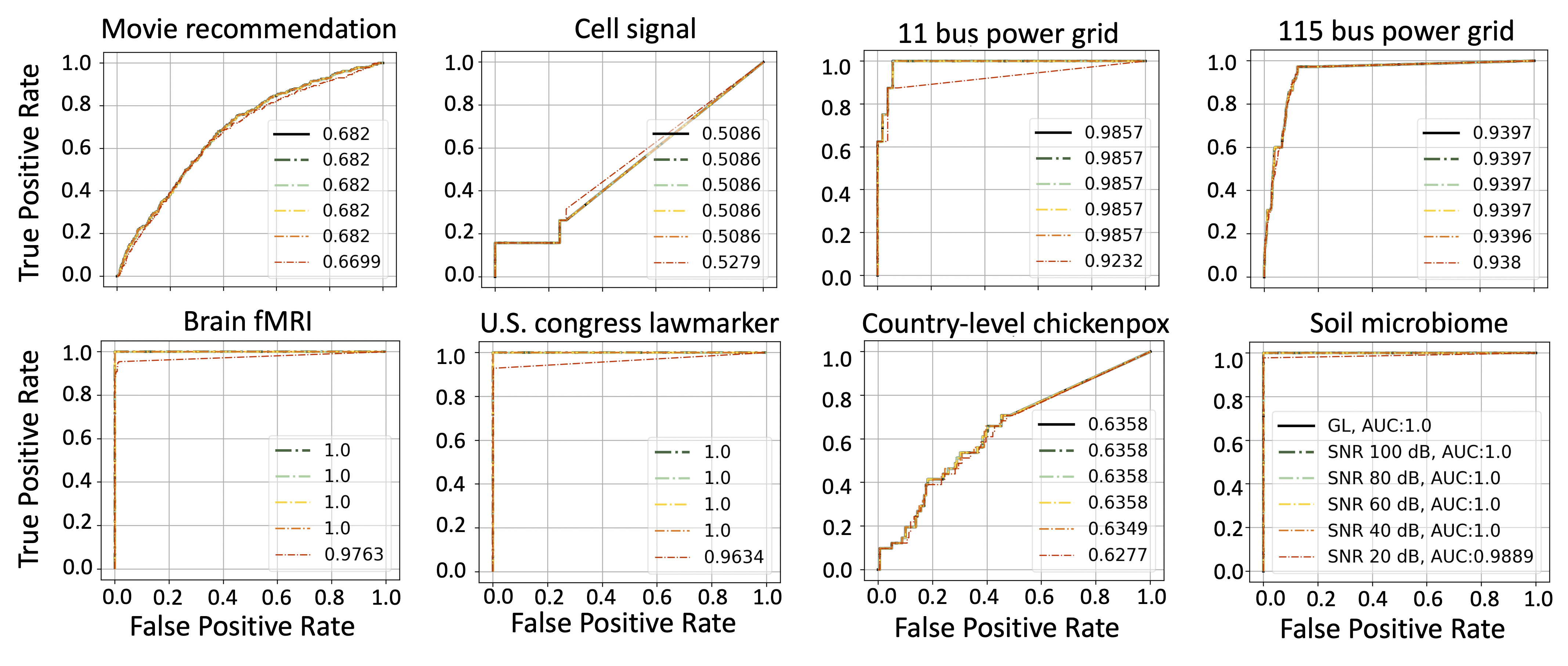}
    \caption{ROC and AUC of our approach and vanilla Graphical Lasso on eight real-world datasets.}
    \label{roc}
    \vspace{-2mm}
\end{figure*}

\begin{figure}[h]
    \centering
    \begin{subfigure}{0.45\textwidth}
        \centering
        \includegraphics[width=\textwidth, trim=4 8 4 4,clip]{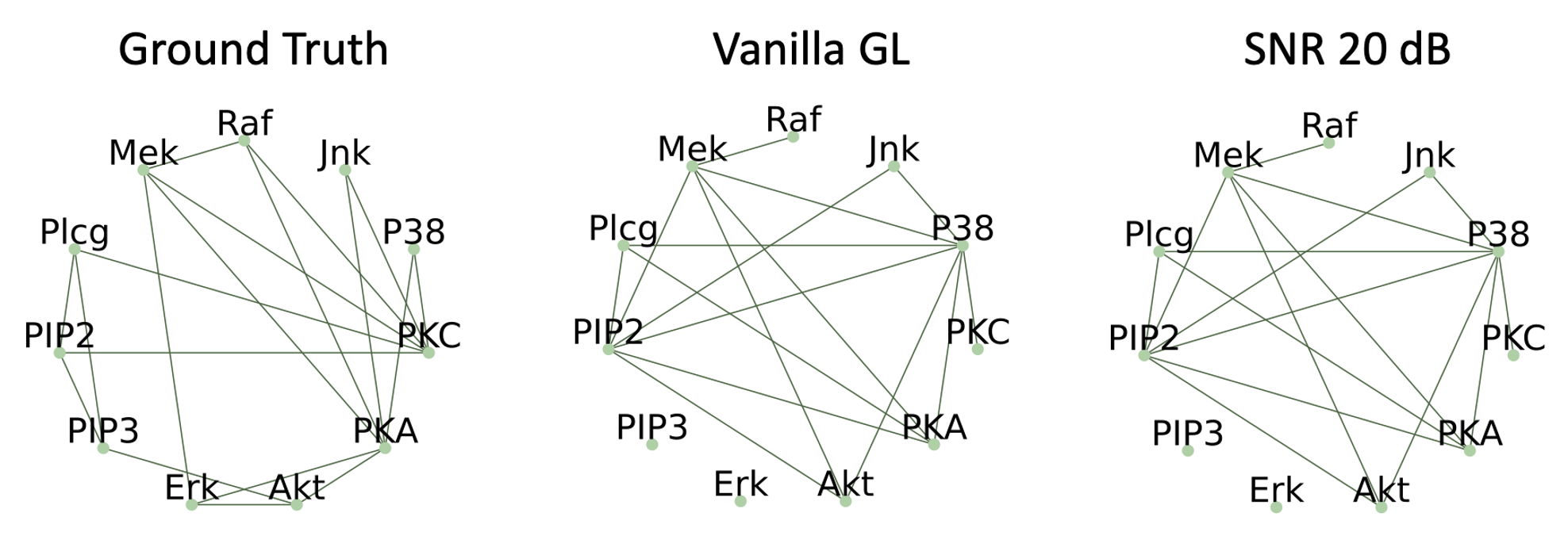}
        \caption{Cell signaling: structure recovery performance.}
        \label{fig:cell_signal_graph}
    \end{subfigure}
    \begin{subfigure}{0.45\textwidth}
        \centering
        \includegraphics[width=\textwidth, trim=4 4 4 4,clip]{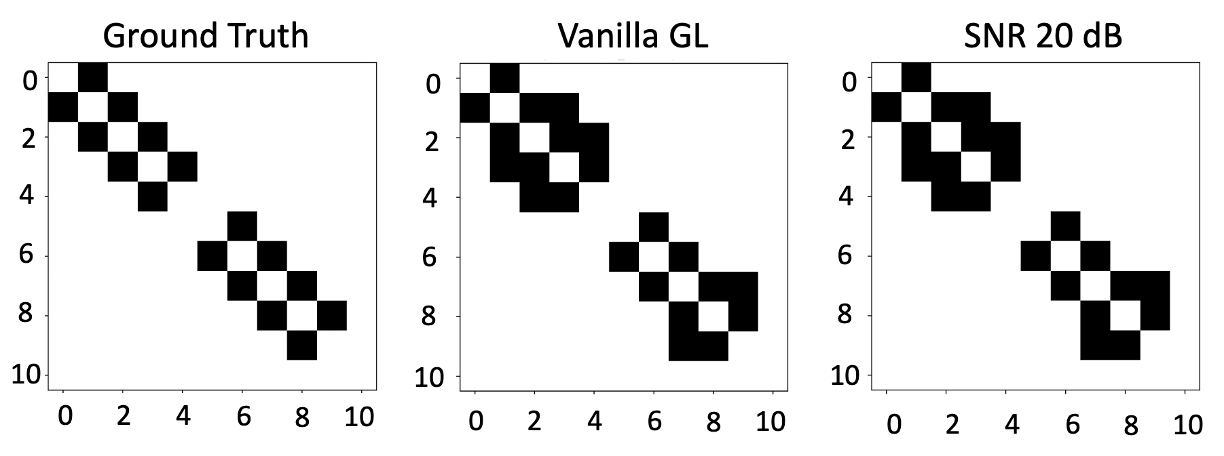}
        \caption{Power system: adjacency matrix performance.}
        \label{fig:power1_graph}
    \end{subfigure}
    \hspace{-2mm}
        \begin{subfigure}{0.45\textwidth}
        \centering
        \includegraphics[width=\textwidth, trim=4 2 4 10,clip]{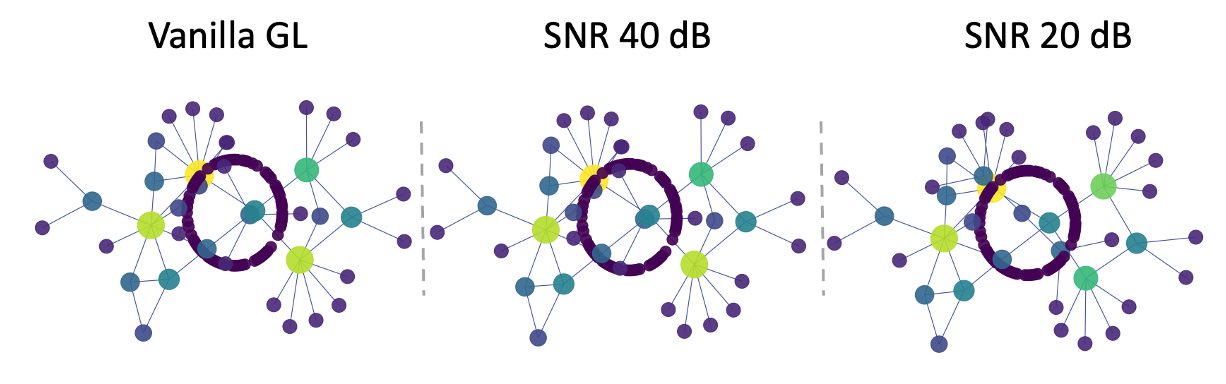}
        \caption{Soil microbiome: structure recovery performance.}
        \label{fig:soil_graph}
    \end{subfigure}
    \begin{subfigure}{0.45\textwidth}
        \centering
        \includegraphics[width=\textwidth, trim=4 4 4 4, clip]{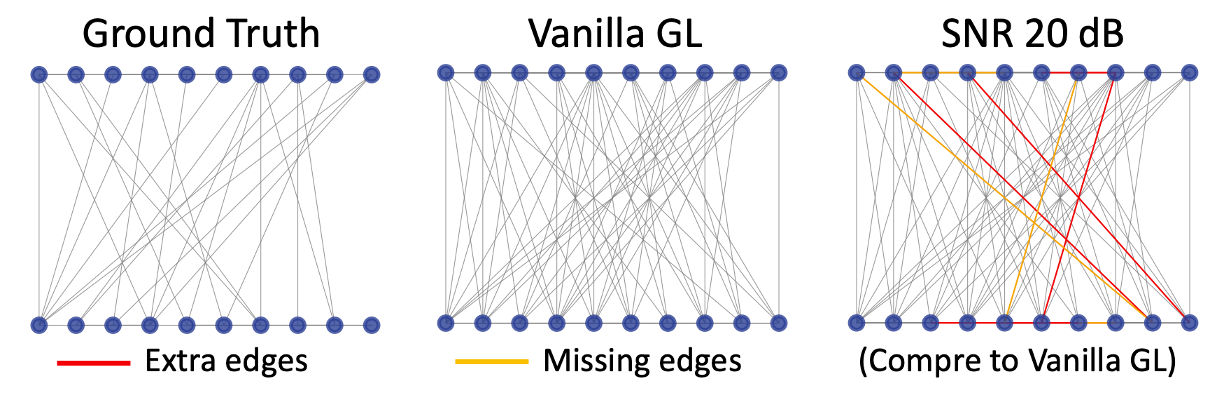}
        \caption{Chickenpox: structure recovery performance.}
        \label{fig:chickenpox_graph_3}
    \end{subfigure}
\caption{Cell signaling, Power system, Chickenpox, and Soil microbiome datasets.}
\label{fig:four datasets}
\vspace{-2mm}
\end{figure}

\begin{figure}[h]
    \centering
    \begin{subfigure}{0.45\textwidth}
        \centering
        \includegraphics[width=\textwidth, trim=8 8 8 8,clip]{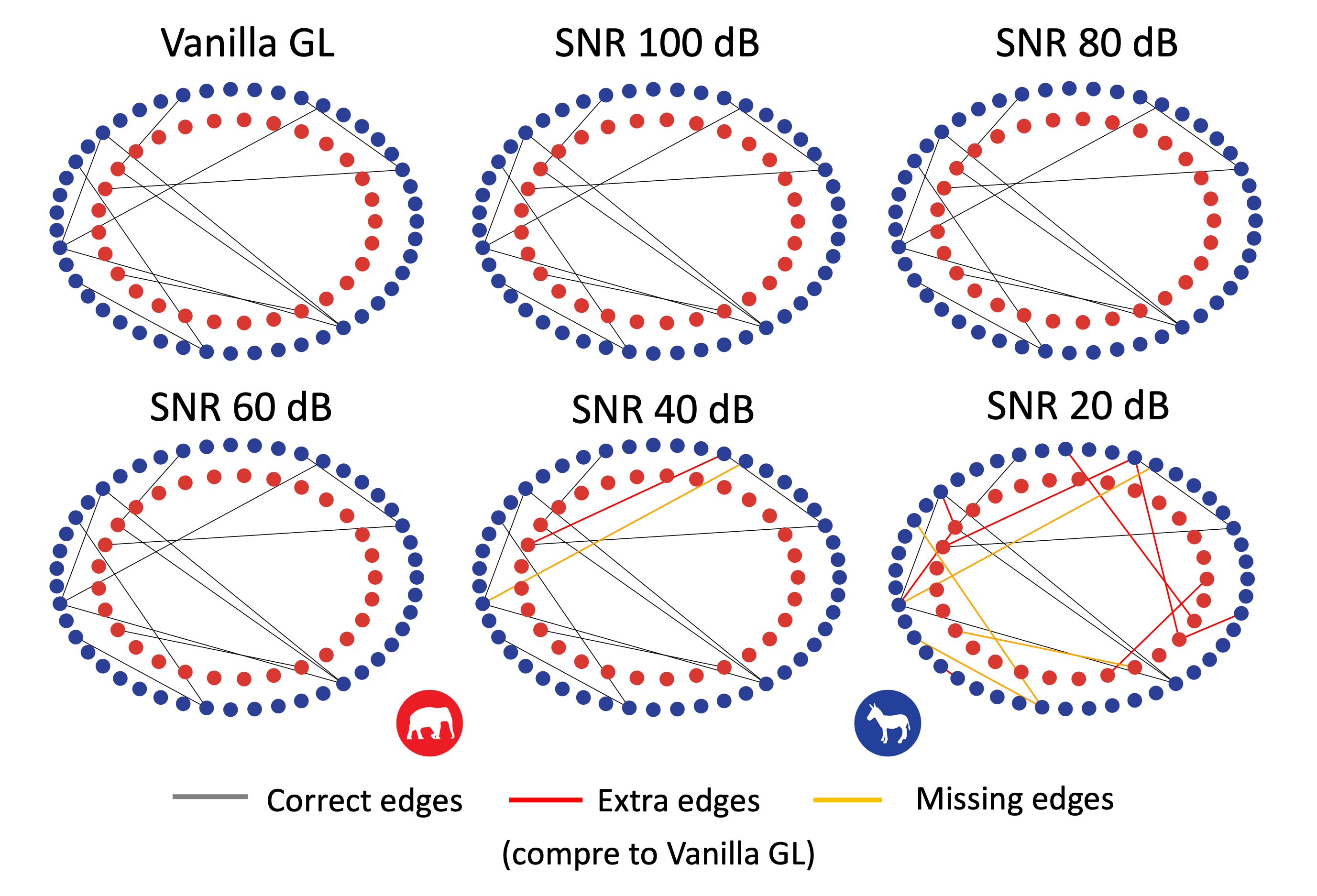}
        \caption{Lawmakers in U.S. Congress: Red and blue represent the Republican and Democratic parties.}
        \label{fig:congress_graph}
    \end{subfigure}
    \begin{subfigure}{0.45\textwidth}
        \centering
        \includegraphics[width=\textwidth, trim=8 8 8 8, clip]{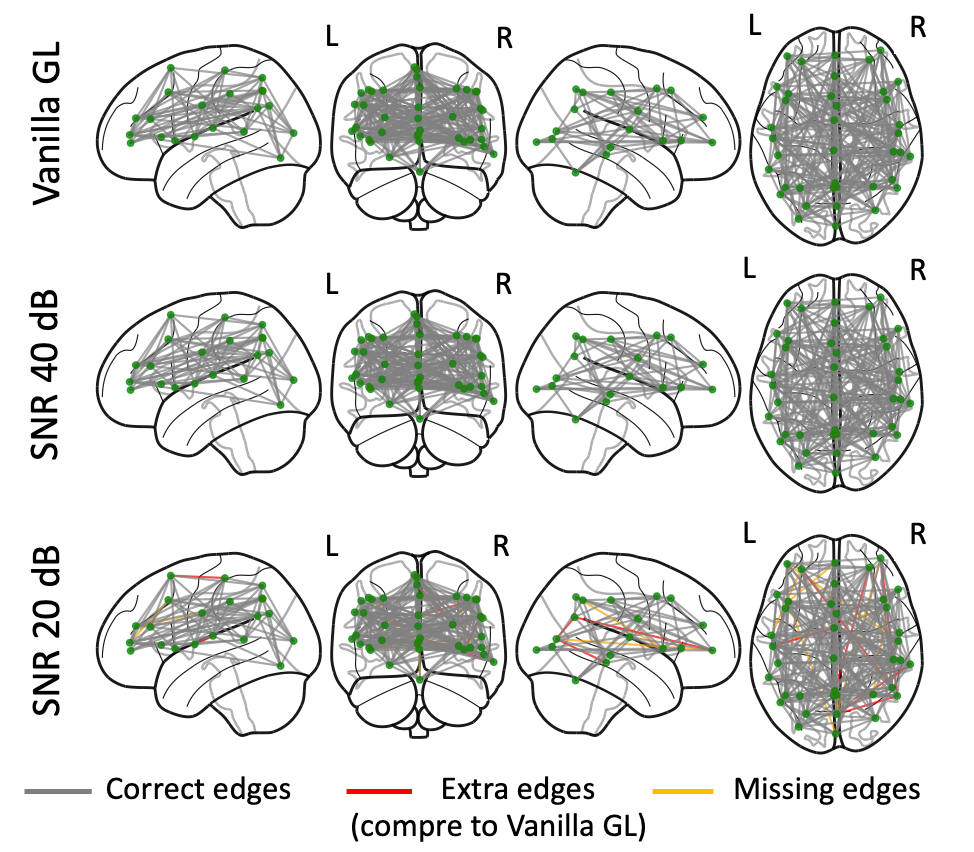}
        \caption{fMRI of Brain: structure recovery.}
        \label{fig:brain_graph}
    \end{subfigure}
\caption{Lawmakers in U.S. Congress, and fMRI of Brain datasets.}
\label{fig:two datasets}
\vspace{-2mm}
\end{figure}

\noindent \textbf{Power Grid.}
We evaluate 11-bus grid data \cite{liao2018urban} to study the stability of a power network’s underlying topology. 
Figure \ref{baselines} compares our approach to three baseline methods: G-Wishart \cite{kuismin2016use}, SCIO \cite{liu2015fast}, and Neighborhood Selection \cite{meinshausen2006high} at 20 dB noise. Our method achieves a visibly closer match to the ground truth. The rightmost plot illustrates the privacy–utility tradeoff, showing that our AUC remains high across a wide range of 
$\epsilon$. 
Including power grids, the ROC and AUC of our approach for all real-world datasets are shown in Table \ref{AUC} and Figure \ref{roc}, and all the visualizations of structure recovery are shown in Figure \ref{fig:four datasets} and \ref{fig:two datasets}. 

\begin{table}[t]
  \caption{AUC and $\epsilon$ of our approach on different datasets with varying noise.}
  \label{AUC}
  \centering
  \resizebox{1\columnwidth}{!}{
  \begin{tabular}{cc|c|ccccc|ccc}
    \toprule[1pt]
    \multicolumn{3}{c|}{} &\multicolumn{5}{c|}{\textbf{Continuous Dataset}}& \multicolumn{3}{c}{\textbf{Discrete Dataset}} \\
    \cmidrule(r){4-11}
    \multicolumn{3}{c|}{Dataset}  & Cell & 11-bus  & 115-bus & Brain & U.S. Congress & Movie & County-level & Soil\\
    \multicolumn{3}{c|}{} & Signalling & Power Grid & Power Grid & fMRI & Lawmakers  & Recommendation & Chickenpox & Microbiome\\
    \midrule
    \multicolumn{3}{c|}{Ground Truth Exists $?$}
    & $\checkmark$  & $\checkmark$ & $\checkmark$ & $\times$ & $\times$ & $\checkmark$ & $\checkmark$ & $\times$ \\
    \midrule[1pt]
    \midrule[1pt]

    
    & \multirow{1}{*}{No Noise} & AUC


    & \makecell{0.71 \\ $\pm1.1\text{e-16}$} 
    & \makecell{0.98 \\ $\pm1.1\text{e-16}$} 
    & \makecell{0.89 \\ $\pm1.1\text{e-16}$}  
    & \makecell{1.0 \\ $\pm0.0$} 
    & \makecell{1.0 \\ $\pm0.0$}  
    & \makecell{0.68 \\ $\pm1.1\text{e-16}$} 
    & \makecell{0.63 \\ $\pm1.1\text{e-16}$}
    & \makecell{1.0 \\ $\pm0.0$}
    \\
    
    \cmidrule[1pt]{2-11}
    
    & \multirow{2}{*}{100 dB} 
    & $\epsilon$ & $ 8.55 \text{e4}$  & $ 5.45 \text{e3} $ & $ 7.48 \text{e3}$  & $  7.18\text{e4}$ & $  6.40\text{e5}$  & $  4.03\text{e2} $ & $  1.99 \text{e6}$ & $  1.34 \text{e7} $ \\
    \cmidrule{3-11}
    & & AUC 
    & \makecell{0.71 \\ $\pm1.1\text{e-16}$}  
    & \makecell{0.98 \\ $\pm1.1\text{e-16}$} 
    & \makecell{0.89 \\ $\pm6.8\text{e-6}$}  
    & \makecell{1.0 \\ $\pm0.0$} 
    & \makecell{1.0 \\ $\pm0.0$} 
    & \makecell{0.68 \\ $\pm1.1\text{e-16}$} 
    & \makecell{0.63 \\ $\pm1.1\text{e-16}$} 
    & \makecell{1.0 \\ $\pm0.0$}
    \\

    \cmidrule[1pt]{2-11}
    
    & \multirow{2}{*}{80 dB} 
    &$\epsilon$ & $  8.55 \text{e3} $  & $  5.45\text{e2}$ & $  7.48\text{e2}$  & $ 7.18\text{e3} $ & $  6.40\text{e4} $  & $   40.32  $ & $  1.99 \text{e5}$ & $  1.34 \text{e6} $ \\
    \cmidrule{3-11}
    & & AUC 
    & \makecell{0.71 \\ $\pm1.1\text{e-16}$}  
    & \makecell{0.98 \\ $\pm1.1\text{e-16}$} 
    & \makecell{0.89 \\ $\pm5.3\text{e-5}$}  
    & \makecell{1.0 \\ $\pm0.0$} 
    & \makecell{1.0 \\ $\pm0.0$} 
    & \makecell{0.68 \\ $\pm1.1\text{e-16}$} 
    & \makecell{0.63 \\ $\pm1.1\text{e-16}$} 
    & \makecell{1.0 \\ $\pm0.0$}
    \\
    \cmidrule[1pt]{2-11}

    & \multirow{2}{*}{60 dB}
    &$\epsilon$ & $  8.55 \text{e2} $ & $  54.50 $ & $  74.84$  & $  7.18\text{e2} $ & $  6.40\text{e3}$  & $  4.03 $ & $  1.99 \text{e4}$ & $  1.34 \text{e5} $ \\
    \cmidrule{3-11}
    && AUC 
    & \makecell{0.71 \\ $\pm1.1\text{e-16}$}  
    & \makecell{0.98 \\ $\pm1.1\text{e-16}$} 
    & \makecell{0.89 \\ $\pm2.1\text{e-3}$}  
    & \makecell{1.0 \\ $\pm5.5\text{e-4}$} 
    & \makecell{0.98 \\ $\pm2.11\text{e-2}$} 
    & \makecell{0.68 \\ $\pm1.1\text{e-16}$} 
    & \makecell{0.63 \\ $\pm1.1\text{e-16}$} 
    & \makecell{1.0 \\ $\pm0.0$}
    \\
    \cmidrule[1pt]{2-11}

    & \multirow{2}{*}{40 dB}
    &$\epsilon$ & $  85.5$  & $  5.45 $ & $  7.48$  & $ 71.79 $ & $  6.40\text{e2} $  & $  0.40 $ & $  1.99 \text{e3} $ & $  1.34 \text{e4}$ \\
    \cmidrule{3-11}
    & & AUC 
    & \makecell{0.71 \\ $\pm1.0\text{e-3}$}  
    & \makecell{0.98 \\ $\pm9.5\text{e-4}$} 
    & \makecell{0.84 \\ $\pm1.8\text{e-2}$}  
    & \makecell{0.99 \\ $\pm1.3\text{e-3}$} 
    & \makecell{0.95 \\ $\pm1.68\text{e-2}$} 
    & \makecell{0.68 \\ $\pm1.1\text{e-16}$} 
    & \makecell{0.63 \\ $\pm2.4\text{e-3}$} 
    & \makecell{1.0 \\ $\pm0.0$}
    \\
    \cmidrule[1pt]{2-11}

    & \multirow{2}{*}{20 dB}
    &$\epsilon$ & $  8.55$  & $  0.54$ & $  0.75$  & $  7.18$ & $  64.04 $  & $  4.03\text{e-2} $ & $  1.99 \text{e2} $ & $ 1.34 \text{e3} $ \\
    \cmidrule{3-11}
    & & AUC 
    & \makecell{0.70 \\ $\pm2.1\text{e-3}$}  
    & \makecell{0.97 \\ $\pm2.9\text{e-2}$} 
    & \makecell{0.73 \\ $\pm2.9\text{e-2}$}  
    & \makecell{0.89 \\ $\pm1.36\text{e-2}$} 
    & \makecell{0.87 \\ $\pm7.34\text{e-2}$} 
    & \makecell{0.67 \\ $\pm1.1\text{e-16}$} 
    & \makecell{0.63 \\ $\pm5.8\text{e-3}$} 
    & \makecell{0.99 \\ $\pm7.1\text{e-3}$}
    \\
    \cmidrule[1pt]{2-11}
    
    & \multirow{2}{*}{10 dB}
    &$\epsilon$ & $  0.85 $  & $  0.05 $ & $  0.07 $ & $  0.72$ & $  6.40$  & $  4.03\text{e-3}  $ & $  19.9 $ & $ 1.34 \text{e2}$ \\
    \cmidrule{3-11}
    & & AUC 
    & \makecell{0.62 \\ $\pm5.8\text{e-2}$}  
    & \makecell{0.85 \\ $\pm9.6\text{e-3}$} 
    & \makecell{0.56 \\ $\pm3.45\text{e-2}$}  
    & \makecell{0.68 \\ $\pm9.6\text{e-3}$} 
    & \makecell{0.50 \\ $\pm2.86\text{e-3}$} 
    & \makecell{0.67 \\ $\pm1.1\text{e-16}$} 
    & \makecell{0.50 \\ $\pm1.1\text{e-2}$} 
    & \makecell{0.75 \\ $\pm0.18$}
    \\
    \bottomrule[1pt]
  \end{tabular}
  }
\end{table}

\label{sec: Applications on Real-World Datasets}



\noindent\textbf{Movie Recommendation (Discrete Dataset).}
Such a task aims to capture the relationships between movies and users \cite{lin2021glimg}. 
But, learning the graph can 
leak private information 
\cite{mcsherry2009differentially}.
We use MovieLens \cite{agarwal2011modeling} for evaluation, which consists of ratings provided by $6,040$ users. 
The movie column in Table \ref{AUC} and the first subplot in Figure \ref{roc} show that our method has the same good performance at different levels of noise for preserving privacy while keeping the topology recovery.

\noindent \textbf{Cell Signaling.}
Cell signaling data \cite{sachs2005causal} is a biological dataset containing $7,466$ cells with flow cytometry measurements of $11$ phosphorylated proteins and phospholipids representing $11$ nodes. This dataset is typically used for recovering the relationship of phosphorylated proteins and phospholipids used in graphical lasso models such as \cite{friedman2008sparse}. 
Figure \ref{fig:cell_signal_graph}
shows that 
our approach successfully reconstructs the structure identically to that from Vanilla GL reported in \cite{friedman2008sparse}. 
It shows small graph size and well-defined correlation signals confer resilience against moderate DP noise up to around 20 dB.

\noindent \textbf{fMRI of Brain.} 
We use MSDL Atlas data \cite{varoquaux2011multi} from Nilearn \cite{Nilearn}, where nodes represent brain regions and edges capture inter-region interactions. Without a strict ground-truth network, we compare our differentially private estimator to the vanilla graphical lasso. At 40 dB SNR, the edge structure is nearly identical to the non-private solution (see Figure~\ref{fig:brain_graph}). Noise starts altering the graph noticeably below 20 dB.


\noindent\textbf{U.S. Congress Lawmakers.}
We apply our method to study the latent connections among legislators using the Legislative Effectiveness Score (LES) \cite{volden2018legislative}. 
%
%
We use LES information for the U.S. Congresses from the 93rd to the 102nd Congress (1973-1991). 
We extract 10-year data because most networks are observed only for relatively short periods. 
Figure \ref{fig:congress_graph} 
shows a consistent ability to accurately recover the structure as the ground truth at privacy levels from $100$ to $60$ dB.

\noindent\textbf{County-level Chickenpox (Discrete Dataset).}
Such dataset includes county-level chickenpox cases \cite{rozemberczki2021chickenpox} in Hungary between $2004$ and $2014$. The underlying graph is static, the vertices are counties, and the edges are neighborhoods. The vertex is the weekly count of chickenpox cases. The dataset consists of $517$ snapshots.
Figure \ref{fig:chickenpox_graph_3} shows that our result is similar to the result of 
vanilla graphical lasso. 

\noindent\textbf{Soil Microbiome (Discrete Dataset).}
Microbial abundance networks are a typical application of graphical lasso \cite{kurtz2019disentangling}. Soil Microbiome contains $89$ soil samples \cite{lauber2009pyrosequencing} from North and South America. 
Each node represents an Operational taxonomic unit (OTU). 
Figure \ref{fig:soil_graph}
shows the recovered structures of soil microbiome on different levels of privacy.

One key observation from Table~\ref{AUC} is that 
sparser graphs (i.e., fewer edges) can often be more robust to differential privacy noise because there are fewer parameters to estimate, and the correlation signals tend to be clearer.
Larger sample sizes also improve the robustness to noise, as seen in datasets like Movie, where abundant user ratings help mitigate the added DP perturbations.

\section{Privacy-Utility Trade-off and Empirical Analysis}
\label{sec:privacy_utility_analysis}

We provide an empirical evaluation of the trade-off between the analytical utility of the recovered graph and the privacy guarantees afforded by our Gaussian privatization mechanism. To quantify this relationship, we analyze the correlation between the SNR, the resulting privacy budget $\epsilon$, and the Area Under the Receiver Operating Characteristic Curve AUC score. The privacy budget $\epsilon$ is derived directly from the noise standard deviation $\sigma_n$ and the data sensitivity $\Delta_f$, where a lower SNR corresponds to a higher noise level and, consequently, a stronger privacy guarantee (lower $\epsilon$).

The results of our trade-off analysis, presented in Table~2 and Figure~3, indicate that our framework maintains robust estimation performance even under rigorous privacy regimes. We observe that at high noise levels, such as 20 dB, the method achieves a graph structure recovery that closely matches the ground truth, outperforming baseline methods like Neighborhood Selection and G-Wishart. While the graph structure begins to show noticeable alterations below 20 dB, the proposed unbiased estimator $\hat{\Theta}$ successfully preserves the essential topology required for analysis across a wide range of privacy budgets. This confirms that the injected Gaussian noise effectively satisfies the Differential Privacy requirements without destroying the statistical signals necessary for sparse inverse covariance estimation.

Furthermore, we evaluate the impact of data properties, specifically sample size and graph sparsity, on the privacy-utility frontier. As shown in our synthetic dataset experiments, larger sample sizes significantly improve robustness to noise, as the statistical consistency of the empirical covariance matrix helps mitigate the impact of the added perturbations. Additionally, we find that sparser graphs—those with fewer edges—are inherently more resilient to differential privacy noise. This is because there are fewer non-zero parameters to estimate, allowing the correlation signals to remain distinct even when the privacy budget $\epsilon$ is tightened. These findings provide data publishers with a practical guide for balancing regulatory compliance (GDPR) with the analytical needs of the research community, particularly for large-scale, sparse datasets.

\section{Conclusion}
\label{sec:conclusion}
We propose a performance-guaranteed, privacy-preserving approach to graph structure learning that ensures robust data protection while maintaining analytical utility. By integrating Gaussian differential privacy, we develop a maximum likelihood estimation framework for sparse graph learning that preserves the privacy of raw data and extends applicability from continuous to discrete datasets. Empirical evaluations on diverse real-world and synthetic datasets demonstrate that the proposed method effectively balances privacy protection and estimation accuracy, making it a viable solution for privacy-conscious graph analysis in modern machine learning. Future work will extend this approach to broader classes of graphical models, including dynamic and temporal networks, and refine privacy-utility tradeoffs through adaptive noise mechanisms.


\bibliographystyle{ACM-Reference-Format}
\bibliography{mybibliography}


\begin{thebibliography}{55}


\ifx \showCODEN    \undefined \def \showCODEN     #1{\unskip}     \fi
\ifx \showDOI      \undefined \def \showDOI       #1{#1}\fi
\ifx \showISBNx    \undefined \def \showISBNx     #1{\unskip}     \fi
\ifx \showISBNxiii \undefined \def \showISBNxiii  #1{\unskip}     \fi
\ifx \showISSN     \undefined \def \showISSN      #1{\unskip}     \fi
\ifx \showLCCN     \undefined \def \showLCCN      #1{\unskip}     \fi
\ifx \shownote     \undefined \def \shownote      #1{#1}          \fi
\ifx \showarticletitle \undefined \def \showarticletitle #1{#1}   \fi
\ifx \showURL      \undefined \def \showURL       {\relax}        \fi
\providecommand\bibfield[2]{#2}
\providecommand\bibinfo[2]{#2}
\providecommand\natexlab[1]{#1}
\providecommand\showeprint[2][]{arXiv:#2}

\bibitem[Abadi et~al\mbox{.}(2016)]%
        {abadi2016deep}
\bibfield{author}{\bibinfo{person}{Martin Abadi}, \bibinfo{person}{Andy Chu}, \bibinfo{person}{Ian Goodfellow}, \bibinfo{person}{H~Brendan McMahan}, \bibinfo{person}{Ilya Mironov}, \bibinfo{person}{Kunal Talwar}, {and} \bibinfo{person}{Li Zhang}.} \bibinfo{year}{2016}\natexlab{}.
\newblock \showarticletitle{Deep learning with differential privacy}. In \bibinfo{booktitle}{\emph{Proceedings of the 2016 ACM SIGSAC conference on computer and communications security}}. \bibinfo{pages}{308--318}.
\newblock


\bibitem[Agarwal et~al\mbox{.}(2011)]%
        {agarwal2011modeling}
\bibfield{author}{\bibinfo{person}{Deepak Agarwal}, \bibinfo{person}{Liang Zhang}, {and} \bibinfo{person}{Rahul Mazumder}.} \bibinfo{year}{2011}\natexlab{}.
\newblock \showarticletitle{Modeling item-item similarities for personalized recommendations on Yahoo! front page}.
\newblock \bibinfo{journal}{\emph{The Annals of applied statistics}} (\bibinfo{year}{2011}), \bibinfo{pages}{1839--1875}.
\newblock


\bibitem[Archana et~al\mbox{.}(2018)]%
        {archana2018study}
\bibfield{author}{\bibinfo{person}{RA Archana}, \bibinfo{person}{Ravindra~S Hegadi}, {and} \bibinfo{person}{TN Manjunath}.} \bibinfo{year}{2018}\natexlab{}.
\newblock \showarticletitle{A study on big data privacy protection models using data masking methods}.
\newblock \bibinfo{journal}{\emph{International Journal of Electrical and Computer Engineering}} \bibinfo{volume}{8}, \bibinfo{number}{5} (\bibinfo{year}{2018}), \bibinfo{pages}{3976}.
\newblock


\bibitem[Auer et~al\mbox{.}(2007)]%
        {auer2007dbpedia}
\bibfield{author}{\bibinfo{person}{S{\"o}ren Auer}, \bibinfo{person}{Christian Bizer}, \bibinfo{person}{Georgi Kobilarov}, \bibinfo{person}{Jens Lehmann}, \bibinfo{person}{Richard Cyganiak}, {and} \bibinfo{person}{Zachary Ives}.} \bibinfo{year}{2007}\natexlab{}.
\newblock \showarticletitle{Dbpedia: A nucleus for a web of open data}. In \bibinfo{booktitle}{\emph{international semantic web conference}}. Springer, \bibinfo{pages}{722--735}.
\newblock


\bibitem[Banerjee et~al\mbox{.}(2008)]%
        {banerjee2008model}
\bibfield{author}{\bibinfo{person}{Onureena Banerjee}, \bibinfo{person}{Laurent El~Ghaoui}, {and} \bibinfo{person}{Alexandre d'Aspremont}.} \bibinfo{year}{2008}\natexlab{}.
\newblock \showarticletitle{Model selection through sparse maximum likelihood estimation for multivariate Gaussian or binary data}.
\newblock \bibinfo{journal}{\emph{The Journal of Machine Learning Research}}  \bibinfo{volume}{9} (\bibinfo{year}{2008}), \bibinfo{pages}{485--516}.
\newblock


\bibitem[Bennett et~al\mbox{.}(2007)]%
        {bennett2007netflix}
\bibfield{author}{\bibinfo{person}{James Bennett}, \bibinfo{person}{Stan Lanning}, {et~al\mbox{.}}} \bibinfo{year}{2007}\natexlab{}.
\newblock \showarticletitle{The netflix prize}. In \bibinfo{booktitle}{\emph{Proceedings of KDD cup and workshop}}, Vol.~\bibinfo{volume}{2007}. New York, \bibinfo{pages}{35}.
\newblock


\bibitem[Boyd et~al\mbox{.}(2011)]%
        {boyd2011distributed}
\bibfield{author}{\bibinfo{person}{Stephen Boyd}, \bibinfo{person}{Neal Parikh}, \bibinfo{person}{Eric Chu}, \bibinfo{person}{Borja Peleato}, \bibinfo{person}{Jonathan Eckstein}, {et~al\mbox{.}}} \bibinfo{year}{2011}\natexlab{}.
\newblock \showarticletitle{Distributed optimization and statistical learning via the alternating direction method of multipliers}.
\newblock \bibinfo{journal}{\emph{Foundations and Trends{\textregistered} in Machine learning}} \bibinfo{volume}{3}, \bibinfo{number}{1} (\bibinfo{year}{2011}), \bibinfo{pages}{1--122}.
\newblock


\bibitem[Brillinger(1996)]%
        {brillinger1996remarks}
\bibfield{author}{\bibinfo{person}{David~R Brillinger}.} \bibinfo{year}{1996}\natexlab{}.
\newblock \showarticletitle{Remarks concerning graphical models for time series and point processes}.
\newblock \bibinfo{journal}{\emph{Brazilian Review of Econometrics}} \bibinfo{volume}{16}, \bibinfo{number}{1} (\bibinfo{year}{1996}), \bibinfo{pages}{1--23}.
\newblock


\bibitem[Canonne et~al\mbox{.}(2020)]%
        {canonne2020discrete}
\bibfield{author}{\bibinfo{person}{Cl{\'e}ment~L Canonne}, \bibinfo{person}{Gautam Kamath}, {and} \bibinfo{person}{Thomas Steinke}.} \bibinfo{year}{2020}\natexlab{}.
\newblock \showarticletitle{The discrete gaussian for differential privacy}.
\newblock \bibinfo{journal}{\emph{Advances in Neural Information Processing Systems}}  \bibinfo{volume}{33} (\bibinfo{year}{2020}), \bibinfo{pages}{15676--15688}.
\newblock


\bibitem[Chaudhuri et~al\mbox{.}(2011)]%
        {chaudhuri2011differentially}
\bibfield{author}{\bibinfo{person}{Kamalika Chaudhuri}, \bibinfo{person}{Claire Monteleoni}, {and} \bibinfo{person}{Anand~D Sarwate}.} \bibinfo{year}{2011}\natexlab{}.
\newblock \showarticletitle{Differentially private empirical risk minimization.}
\newblock \bibinfo{journal}{\emph{Journal of Machine Learning Research}} \bibinfo{volume}{12}, \bibinfo{number}{3} (\bibinfo{year}{2011}).
\newblock


\bibitem[contributors({[n.\,d.]})]%
        {Nilearn}
\bibfield{author}{\bibinfo{person}{Nilearn contributors}.} \bibinfo{year}{[n.\,d.]}\natexlab{}.
\newblock \bibinfo{booktitle}{\emph{{nilearn}}}.
\newblock
\urldef\tempurl%
\url{https://doi.org/10.5281/zenodo.8397156}
\showDOI{\tempurl}


\bibitem[Cui and Lee(2020)]%
        {cui2020coaid}
\bibfield{author}{\bibinfo{person}{Limeng Cui} {and} \bibinfo{person}{Dongwon Lee}.} \bibinfo{year}{2020}\natexlab{}.
\newblock \showarticletitle{Coaid: Covid-19 healthcare misinformation dataset}.
\newblock \bibinfo{journal}{\emph{arXiv preprint arXiv:2006.00885}} (\bibinfo{year}{2020}).
\newblock


\bibitem[Dempster(1972)]%
        {dempster1972covariance}
\bibfield{author}{\bibinfo{person}{Arthur~P Dempster}.} \bibinfo{year}{1972}\natexlab{}.
\newblock \showarticletitle{Covariance selection}.
\newblock \bibinfo{journal}{\emph{Biometrics}} (\bibinfo{year}{1972}), \bibinfo{pages}{157--175}.
\newblock


\bibitem[Deng et~al\mbox{.}(2009)]%
        {deng2009imagenet}
\bibfield{author}{\bibinfo{person}{Jia Deng}, \bibinfo{person}{Wei Dong}, \bibinfo{person}{Richard Socher}, \bibinfo{person}{Li-Jia Li}, \bibinfo{person}{Kai Li}, {and} \bibinfo{person}{Li Fei-Fei}.} \bibinfo{year}{2009}\natexlab{}.
\newblock \showarticletitle{Imagenet: A large-scale hierarchical image database}. In \bibinfo{booktitle}{\emph{2009 IEEE conference on computer vision and pattern recognition}}. Ieee, \bibinfo{pages}{248--255}.
\newblock


\bibitem[Domingo-Ferrer and Soria-Comas(2015)]%
        {domingo2015t}
\bibfield{author}{\bibinfo{person}{Josep Domingo-Ferrer} {and} \bibinfo{person}{Jordi Soria-Comas}.} \bibinfo{year}{2015}\natexlab{}.
\newblock \showarticletitle{From t-closeness to differential privacy and vice versa in data anonymization}.
\newblock \bibinfo{journal}{\emph{Knowledge-Based Systems}}  \bibinfo{volume}{74} (\bibinfo{year}{2015}), \bibinfo{pages}{151--158}.
\newblock


\bibitem[Dong et~al\mbox{.}(2019)]%
        {dong2019gaussian}
\bibfield{author}{\bibinfo{person}{Jinshuo Dong}, \bibinfo{person}{Aaron Roth}, {and} \bibinfo{person}{Weijie~J Su}.} \bibinfo{year}{2019}\natexlab{}.
\newblock \showarticletitle{Gaussian differential privacy}.
\newblock \bibinfo{journal}{\emph{arXiv preprint arXiv:1905.02383}} (\bibinfo{year}{2019}).
\newblock


\bibitem[Dwork et~al\mbox{.}(2006)]%
        {dwork2006calibrating}
\bibfield{author}{\bibinfo{person}{Cynthia Dwork}, \bibinfo{person}{Frank McSherry}, \bibinfo{person}{Kobbi Nissim}, {and} \bibinfo{person}{Adam Smith}.} \bibinfo{year}{2006}\natexlab{}.
\newblock \showarticletitle{Calibrating noise to sensitivity in private data analysis}. In \bibinfo{booktitle}{\emph{Theory of Cryptography: Third Theory of Cryptography Conference, TCC 2006, New York, NY, USA, March 4-7, 2006. Proceedings 3}}. Springer, \bibinfo{pages}{265--284}.
\newblock


\bibitem[Erlingsson et~al\mbox{.}(2014)]%
        {erlingsson2014rappor}
\bibfield{author}{\bibinfo{person}{{\'U}lfar Erlingsson}, \bibinfo{person}{Vasyl Pihur}, {and} \bibinfo{person}{Aleksandra Korolova}.} \bibinfo{year}{2014}\natexlab{}.
\newblock \showarticletitle{Rappor: Randomized aggregatable privacy-preserving ordinal response}. In \bibinfo{booktitle}{\emph{Proceedings of the 2014 ACM SIGSAC conference on computer and communications security}}. \bibinfo{pages}{1054--1067}.
\newblock


\bibitem[Friedman et~al\mbox{.}(2008)]%
        {friedman2008sparse}
\bibfield{author}{\bibinfo{person}{Jerome Friedman}, \bibinfo{person}{Trevor Hastie}, {and} \bibinfo{person}{Robert Tibshirani}.} \bibinfo{year}{2008}\natexlab{}.
\newblock \showarticletitle{Sparse inverse covariance estimation with the graphical lasso}.
\newblock \bibinfo{journal}{\emph{Biostatistics}} \bibinfo{volume}{9}, \bibinfo{number}{3} (\bibinfo{year}{2008}), \bibinfo{pages}{432--441}.
\newblock


\bibitem[Fung et~al\mbox{.}(2010)]%
        {fung2010privacy}
\bibfield{author}{\bibinfo{person}{Benjamin~CM Fung}, \bibinfo{person}{Ke Wang}, \bibinfo{person}{Rui Chen}, {and} \bibinfo{person}{Philip~S Yu}.} \bibinfo{year}{2010}\natexlab{}.
\newblock \showarticletitle{Privacy-preserving data publishing: A survey of recent developments}.
\newblock \bibinfo{journal}{\emph{ACM Computing Surveys (Csur)}} \bibinfo{volume}{42}, \bibinfo{number}{4} (\bibinfo{year}{2010}), \bibinfo{pages}{1--53}.
\newblock


\bibitem[Guo et~al\mbox{.}(2023)]%
        {guo2023graph}
\bibfield{author}{\bibinfo{person}{Muhao Guo}, \bibinfo{person}{Qiushi Cui}, {and} \bibinfo{person}{Yang Weng}.} \bibinfo{year}{2023}\natexlab{}.
\newblock \showarticletitle{Graph mining for classifying and localizing solar panels in distribution grids}. In \bibinfo{booktitle}{\emph{2023 Panda Forum on Power and Energy (PandaFPE)}}. IEEE, \bibinfo{pages}{1743--1747}.
\newblock


\bibitem[Haney et~al\mbox{.}(2017)]%
        {haney2017utility}
\bibfield{author}{\bibinfo{person}{Samuel Haney}, \bibinfo{person}{Ashwin Machanavajjhala}, \bibinfo{person}{John~M Abowd}, \bibinfo{person}{Matthew Graham}, \bibinfo{person}{Mark Kutzbach}, {and} \bibinfo{person}{Lars Vilhuber}.} \bibinfo{year}{2017}\natexlab{}.
\newblock \showarticletitle{Utility cost of formal privacy for releasing national employer-employee statistics}. In \bibinfo{booktitle}{\emph{Proceedings of the 2017 ACM International Conference on Management of Data}}. \bibinfo{pages}{1339--1354}.
\newblock


\bibitem[Hastie et~al\mbox{.}(2015)]%
        {10.5555/2834535}
\bibfield{author}{\bibinfo{person}{Trevor Hastie}, \bibinfo{person}{Robert Tibshirani}, {and} \bibinfo{person}{Martin Wainwright}.} \bibinfo{year}{2015}\natexlab{}.
\newblock \bibinfo{booktitle}{\emph{Statistical Learning with Sparsity: The Lasso and Generalizations}}.
\newblock \bibinfo{publisher}{Chapman \& Hall/CRC}.
\newblock
\showISBNx{1498712169}


\bibitem[Kuismin and Sillanp{\"a}{\"a}(2016)]%
        {kuismin2016use}
\bibfield{author}{\bibinfo{person}{Markku Kuismin} {and} \bibinfo{person}{Mikko~J Sillanp{\"a}{\"a}}.} \bibinfo{year}{2016}\natexlab{}.
\newblock \showarticletitle{Use of Wishart prior and simple extensions for sparse precision matrix estimation}.
\newblock \bibinfo{journal}{\emph{PloS one}} \bibinfo{volume}{11}, \bibinfo{number}{2} (\bibinfo{year}{2016}), \bibinfo{pages}{e0148171}.
\newblock


\bibitem[Kurtz et~al\mbox{.}(2019)]%
        {kurtz2019disentangling}
\bibfield{author}{\bibinfo{person}{Zachary~D Kurtz}, \bibinfo{person}{Richard Bonneau}, {and} \bibinfo{person}{Christian~L M{\"u}ller}.} \bibinfo{year}{2019}\natexlab{}.
\newblock \showarticletitle{Disentangling microbial associations from hidden environmental and technical factors via latent graphical models}.
\newblock \bibinfo{journal}{\emph{BioRxiv}} (\bibinfo{year}{2019}), \bibinfo{pages}{2019--12}.
\newblock


\bibitem[Lauber et~al\mbox{.}(2009)]%
        {lauber2009pyrosequencing}
\bibfield{author}{\bibinfo{person}{Christian~L Lauber}, \bibinfo{person}{Micah Hamady}, \bibinfo{person}{Rob Knight}, {and} \bibinfo{person}{Noah Fierer}.} \bibinfo{year}{2009}\natexlab{}.
\newblock \showarticletitle{Pyrosequencing-based assessment of soil pH as a predictor of soil bacterial community structure at the continental scale}.
\newblock \bibinfo{journal}{\emph{Applied and environmental microbiology}} \bibinfo{volume}{75}, \bibinfo{number}{15} (\bibinfo{year}{2009}), \bibinfo{pages}{5111--5120}.
\newblock


\bibitem[Lauritzen(1996)]%
        {lauritzen1996graphical}
\bibfield{author}{\bibinfo{person}{Steffen~L Lauritzen}.} \bibinfo{year}{1996}\natexlab{}.
\newblock \bibinfo{booktitle}{\emph{Graphical models}}. Vol.~\bibinfo{volume}{17}.
\newblock \bibinfo{publisher}{Clarendon Press}.
\newblock


\bibitem[Li and Wang(2023)]%
        {li2023differentially}
\bibfield{author}{\bibinfo{person}{Huimin Li} {and} \bibinfo{person}{Jinru Wang}.} \bibinfo{year}{2023}\natexlab{}.
\newblock \showarticletitle{Differentially Private Sparse Covariance Matrix Estimation under Lower-Bounded Moment Assumption}.
\newblock \bibinfo{journal}{\emph{Mathematics}} \bibinfo{volume}{11}, \bibinfo{number}{17} (\bibinfo{year}{2023}), \bibinfo{pages}{3670}.
\newblock


\bibitem[Li et~al\mbox{.}(2006)]%
        {li2006t}
\bibfield{author}{\bibinfo{person}{Ninghui Li}, \bibinfo{person}{Tiancheng Li}, {and} \bibinfo{person}{Suresh Venkatasubramanian}.} \bibinfo{year}{2006}\natexlab{}.
\newblock \showarticletitle{t-closeness: Privacy beyond k-anonymity and l-diversity}. In \bibinfo{booktitle}{\emph{2007 IEEE 23rd international conference on data engineering}}. IEEE, \bibinfo{pages}{106--115}.
\newblock


\bibitem[Liao et~al\mbox{.}(2018)]%
        {liao2018urban}
\bibfield{author}{\bibinfo{person}{Yizheng Liao}, \bibinfo{person}{Yang Weng}, \bibinfo{person}{Guangyi Liu}, {and} \bibinfo{person}{Ram Rajagopal}.} \bibinfo{year}{2018}\natexlab{}.
\newblock \showarticletitle{Urban MV and LV distribution grid topology estimation via group lasso}.
\newblock \bibinfo{journal}{\emph{IEEE Transactions on Power Systems}} \bibinfo{volume}{34}, \bibinfo{number}{1} (\bibinfo{year}{2018}), \bibinfo{pages}{12--27}.
\newblock


\bibitem[Lin et~al\mbox{.}(2021)]%
        {lin2021glimg}
\bibfield{author}{\bibinfo{person}{Zhuoyi Lin}, \bibinfo{person}{Lei Feng}, \bibinfo{person}{Rui Yin}, \bibinfo{person}{Chi Xu}, {and} \bibinfo{person}{Chee~Keong Kwoh}.} \bibinfo{year}{2021}\natexlab{}.
\newblock \showarticletitle{GLIMG: Global and local item graphs for top-N recommender systems}.
\newblock \bibinfo{journal}{\emph{Information Sciences}}  \bibinfo{volume}{580} (\bibinfo{year}{2021}), \bibinfo{pages}{1--14}.
\newblock


\bibitem[Liu and Luo(2015)]%
        {liu2015fast}
\bibfield{author}{\bibinfo{person}{Weidong Liu} {and} \bibinfo{person}{Xi Luo}.} \bibinfo{year}{2015}\natexlab{}.
\newblock \showarticletitle{Fast and adaptive sparse precision matrix estimation in high dimensions}.
\newblock \bibinfo{journal}{\emph{Journal of multivariate analysis}}  \bibinfo{volume}{135} (\bibinfo{year}{2015}), \bibinfo{pages}{153--162}.
\newblock


\bibitem[Machanavajjhala et~al\mbox{.}(2007)]%
        {machanavajjhala2007diversity}
\bibfield{author}{\bibinfo{person}{Ashwin Machanavajjhala}, \bibinfo{person}{Daniel Kifer}, \bibinfo{person}{Johannes Gehrke}, {and} \bibinfo{person}{Muthuramakrishnan Venkitasubramaniam}.} \bibinfo{year}{2007}\natexlab{}.
\newblock \showarticletitle{l-diversity: Privacy beyond k-anonymity}.
\newblock \bibinfo{journal}{\emph{Acm transactions on knowledge discovery from data (tkdd)}} \bibinfo{volume}{1}, \bibinfo{number}{1} (\bibinfo{year}{2007}), \bibinfo{pages}{3--es}.
\newblock


\bibitem[Matatov et~al\mbox{.}(2010)]%
        {matatov2010privacy}
\bibfield{author}{\bibinfo{person}{Nissim Matatov}, \bibinfo{person}{Lior Rokach}, {and} \bibinfo{person}{Oded Maimon}.} \bibinfo{year}{2010}\natexlab{}.
\newblock \showarticletitle{Privacy-preserving data mining: A feature set partitioning approach}.
\newblock \bibinfo{journal}{\emph{Information Sciences}} \bibinfo{volume}{180}, \bibinfo{number}{14} (\bibinfo{year}{2010}), \bibinfo{pages}{2696--2720}.
\newblock


\bibitem[McKenna et~al\mbox{.}(2022)]%
        {mckenna2022aim}
\bibfield{author}{\bibinfo{person}{Ryan McKenna}, \bibinfo{person}{Brett Mullins}, \bibinfo{person}{Daniel Sheldon}, {and} \bibinfo{person}{Gerome Miklau}.} \bibinfo{year}{2022}\natexlab{}.
\newblock \showarticletitle{AIM: An Adaptive and Iterative Mechanism for Differentially Private Synthetic Data. CoRR abs/2201.12677 (2022)}.
\newblock \bibinfo{journal}{\emph{arXiv preprint arXiv:2201.12677}} (\bibinfo{year}{2022}).
\newblock


\bibitem[McSherry and Mironov(2009)]%
        {mcsherry2009differentially}
\bibfield{author}{\bibinfo{person}{Frank McSherry} {and} \bibinfo{person}{Ilya Mironov}.} \bibinfo{year}{2009}\natexlab{}.
\newblock \showarticletitle{Differentially private recommender systems: Building privacy into the netflix prize contenders}. In \bibinfo{booktitle}{\emph{Proceedings of the 15th ACM SIGKDD international conference on Knowledge discovery and data mining}}. \bibinfo{pages}{627--636}.
\newblock


\bibitem[Meinshausen and B{\"u}hlmann(2006)]%
        {meinshausen2006high}
\bibfield{author}{\bibinfo{person}{Nicolai Meinshausen} {and} \bibinfo{person}{Peter B{\"u}hlmann}.} \bibinfo{year}{2006}\natexlab{}.
\newblock \showarticletitle{High-dimensional graphs and variable selection with the lasso}.
\newblock  (\bibinfo{year}{2006}).
\newblock


\bibitem[Mivule(2013)]%
        {mivule2013utilizing}
\bibfield{author}{\bibinfo{person}{Kato Mivule}.} \bibinfo{year}{2013}\natexlab{}.
\newblock \showarticletitle{Utilizing noise addition for data privacy, an overview}.
\newblock \bibinfo{journal}{\emph{arXiv preprint arXiv:1309.3958}} (\bibinfo{year}{2013}).
\newblock


\bibitem[Mohammed et~al\mbox{.}(2011)]%
        {mohammed2011differentially}
\bibfield{author}{\bibinfo{person}{Noman Mohammed}, \bibinfo{person}{Rui Chen}, \bibinfo{person}{Benjamin~CM Fung}, {and} \bibinfo{person}{Philip~S Yu}.} \bibinfo{year}{2011}\natexlab{}.
\newblock \showarticletitle{Differentially private data release for data mining}. In \bibinfo{booktitle}{\emph{Proceedings of the 17th ACM SIGKDD international conference on Knowledge discovery and data mining}}. \bibinfo{pages}{493--501}.
\newblock


\bibitem[Narayanan and Shmatikov(2008)]%
        {narayanan2008robust}
\bibfield{author}{\bibinfo{person}{Arvind Narayanan} {and} \bibinfo{person}{Vitaly Shmatikov}.} \bibinfo{year}{2008}\natexlab{}.
\newblock \showarticletitle{Robust de-anonymization of large sparse datasets}. In \bibinfo{booktitle}{\emph{2008 IEEE Symposium on Security and Privacy (sp 2008)}}. IEEE, \bibinfo{pages}{111--125}.
\newblock


\bibitem[Peikert et~al\mbox{.}(2016)]%
        {peikert2016decade}
\bibfield{author}{\bibinfo{person}{Chris Peikert} {et~al\mbox{.}}} \bibinfo{year}{2016}\natexlab{}.
\newblock \showarticletitle{A decade of lattice cryptography}.
\newblock \bibinfo{journal}{\emph{Foundations and trends{\textregistered} in theoretical computer science}} \bibinfo{volume}{10}, \bibinfo{number}{4} (\bibinfo{year}{2016}), \bibinfo{pages}{283--424}.
\newblock


\bibitem[Qiu et~al\mbox{.}(2015)]%
        {qiu2015predicting}
\bibfield{author}{\bibinfo{person}{Jiangtao Qiu}, \bibinfo{person}{Zhangxi Lin}, {and} \bibinfo{person}{Yinghong Li}.} \bibinfo{year}{2015}\natexlab{}.
\newblock \showarticletitle{Predicting customer purchase behavior in the e-commerce context}.
\newblock \bibinfo{journal}{\emph{Electronic commerce research}}  \bibinfo{volume}{15} (\bibinfo{year}{2015}), \bibinfo{pages}{427--452}.
\newblock


\bibitem[Rozemberczki et~al\mbox{.}(2021)]%
        {rozemberczki2021chickenpox}
\bibfield{author}{\bibinfo{person}{Benedek Rozemberczki}, \bibinfo{person}{Paul Scherer}, \bibinfo{person}{Oliver Kiss}, \bibinfo{person}{Rik Sarkar}, {and} \bibinfo{person}{Tamas Ferenci}.} \bibinfo{year}{2021}\natexlab{}.
\newblock \showarticletitle{Chickenpox cases in Hungary: a benchmark dataset for spatiotemporal signal processing with graph neural networks}.
\newblock \bibinfo{journal}{\emph{arXiv preprint arXiv:2102.08100}} (\bibinfo{year}{2021}).
\newblock


\bibitem[Sachs et~al\mbox{.}(2005)]%
        {sachs2005causal}
\bibfield{author}{\bibinfo{person}{Karen Sachs}, \bibinfo{person}{Omar Perez}, \bibinfo{person}{Dana Pe'er}, \bibinfo{person}{Douglas~A Lauffenburger}, {and} \bibinfo{person}{Garry~P Nolan}.} \bibinfo{year}{2005}\natexlab{}.
\newblock \showarticletitle{Causal protein-signaling networks derived from multiparameter single-cell data}.
\newblock \bibinfo{journal}{\emph{Science}} \bibinfo{volume}{308}, \bibinfo{number}{5721} (\bibinfo{year}{2005}), \bibinfo{pages}{523--529}.
\newblock


\bibitem[Shokri and Shmatikov(2015)]%
        {shokri2015privacy}
\bibfield{author}{\bibinfo{person}{Reza Shokri} {and} \bibinfo{person}{Vitaly Shmatikov}.} \bibinfo{year}{2015}\natexlab{}.
\newblock \showarticletitle{Privacy-preserving deep learning}. In \bibinfo{booktitle}{\emph{Proceedings of the 22nd ACM SIGSAC conference on computer and communications security}}. \bibinfo{pages}{1310--1321}.
\newblock


\bibitem[Shrivastava et~al\mbox{.}(2019)]%
        {shrivastava2019glad}
\bibfield{author}{\bibinfo{person}{Harsh Shrivastava}, \bibinfo{person}{Xinshi Chen}, \bibinfo{person}{Binghong Chen}, \bibinfo{person}{Guanghui Lan}, \bibinfo{person}{Srinvas Aluru}, \bibinfo{person}{Han Liu}, {and} \bibinfo{person}{Le Song}.} \bibinfo{year}{2019}\natexlab{}.
\newblock \showarticletitle{GLAD: Learning sparse graph recovery}.
\newblock \bibinfo{journal}{\emph{arXiv preprint arXiv:1906.00271}} (\bibinfo{year}{2019}).
\newblock


\bibitem[Sweeney(2002)]%
        {sweeney2002k}
\bibfield{author}{\bibinfo{person}{Latanya Sweeney}.} \bibinfo{year}{2002}\natexlab{}.
\newblock \showarticletitle{k-anonymity: A model for protecting privacy}.
\newblock \bibinfo{journal}{\emph{International journal of uncertainty, fuzziness and knowledge-based systems}} \bibinfo{volume}{10}, \bibinfo{number}{05} (\bibinfo{year}{2002}), \bibinfo{pages}{557--570}.
\newblock


\bibitem[Touvron et~al\mbox{.}(2023)]%
        {touvron2023llama}
\bibfield{author}{\bibinfo{person}{Hugo Touvron}, \bibinfo{person}{Louis Martin}, \bibinfo{person}{Kevin Stone}, \bibinfo{person}{Peter Albert}, \bibinfo{person}{Amjad Almahairi}, \bibinfo{person}{Yasmine Babaei}, \bibinfo{person}{Nikolay Bashlykov}, \bibinfo{person}{Soumya Batra}, \bibinfo{person}{Prajjwal Bhargava}, \bibinfo{person}{Shruti Bhosale}, {et~al\mbox{.}}} \bibinfo{year}{2023}\natexlab{}.
\newblock \showarticletitle{Llama 2: Open foundation and fine-tuned chat models}.
\newblock \bibinfo{journal}{\emph{arXiv preprint arXiv:2307.09288}} (\bibinfo{year}{2023}).
\newblock


\bibitem[Varoquaux et~al\mbox{.}(2011)]%
        {varoquaux2011multi}
\bibfield{author}{\bibinfo{person}{Ga{\"e}l Varoquaux}, \bibinfo{person}{Alexandre Gramfort}, \bibinfo{person}{Fabian Pedregosa}, \bibinfo{person}{Vincent Michel}, {and} \bibinfo{person}{Bertrand Thirion}.} \bibinfo{year}{2011}\natexlab{}.
\newblock \showarticletitle{Multi-subject dictionary learning to segment an atlas of brain spontaneous activity}. In \bibinfo{booktitle}{\emph{Information Processing in Medical Imaging: 22nd International Conference, IPMI 2011, Kloster Irsee, Germany, July 3-8, 2011. Proceedings 22}}. Springer, \bibinfo{pages}{562--573}.
\newblock


\bibitem[Voigt and Von~dem Bussche(2017)]%
        {voigt2017eu}
\bibfield{author}{\bibinfo{person}{Paul Voigt} {and} \bibinfo{person}{Axel Von~dem Bussche}.} \bibinfo{year}{2017}\natexlab{}.
\newblock \showarticletitle{The eu general data protection regulation ({gdpr})}.
\newblock \bibinfo{journal}{\emph{A Practical Guide, 1st Ed., Cham: Springer International Publishing}} \bibinfo{volume}{10}, \bibinfo{number}{3152676} (\bibinfo{year}{2017}), \bibinfo{pages}{10--5555}.
\newblock


\bibitem[Volden and Wiseman(2018)]%
        {volden2018legislative}
\bibfield{author}{\bibinfo{person}{Craig Volden} {and} \bibinfo{person}{Alan~E Wiseman}.} \bibinfo{year}{2018}\natexlab{}.
\newblock \showarticletitle{Legislative effectiveness in the United States senate}.
\newblock \bibinfo{journal}{\emph{The Journal of Politics}} \bibinfo{volume}{80}, \bibinfo{number}{2} (\bibinfo{year}{2018}), \bibinfo{pages}{731--735}.
\newblock


\bibitem[Wang et~al\mbox{.}(2018)]%
        {wang2018differentially}
\bibfield{author}{\bibinfo{person}{Di Wang}, \bibinfo{person}{Mengdi Huai}, {and} \bibinfo{person}{Jinhui Xu}.} \bibinfo{year}{2018}\natexlab{}.
\newblock \showarticletitle{Differentially private sparse inverse covariance estimation}. In \bibinfo{booktitle}{\emph{2018 IEEE Global Conference on Signal and Information Processing (GlobalSIP)}}. IEEE, \bibinfo{pages}{1139--1143}.
\newblock


\bibitem[Wu et~al\mbox{.}(2019)]%
        {wu2019distributed}
\bibfield{author}{\bibinfo{person}{Chao Wu}, \bibinfo{person}{Fengda Zhang}, {and} \bibinfo{person}{Fei Wu}.} \bibinfo{year}{2019}\natexlab{}.
\newblock \showarticletitle{Distributed modelling approaches for data privacy preserving}. In \bibinfo{booktitle}{\emph{2019 IEEE Fifth International Conference on Multimedia Big Data (BigMM)}}. IEEE, \bibinfo{pages}{357--365}.
\newblock


\bibitem[Yin et~al\mbox{.}(2020)]%
        {yin2020gaussian}
\bibfield{author}{\bibinfo{person}{Hang Yin}, \bibinfo{person}{Xinyue Liu}, {and} \bibinfo{person}{Xiangnan Kong}.} \bibinfo{year}{2020}\natexlab{}.
\newblock \showarticletitle{Gaussian mixture graphical lasso with application to edge detection in brain networks}. In \bibinfo{booktitle}{\emph{2020 IEEE International Conference on Big Data (Big Data)}}. IEEE, \bibinfo{pages}{1430--1435}.
\newblock


\bibitem[Yuan and Lin(2007)]%
        {yuan2007model}
\bibfield{author}{\bibinfo{person}{Ming Yuan} {and} \bibinfo{person}{Yi Lin}.} \bibinfo{year}{2007}\natexlab{}.
\newblock \showarticletitle{Model selection and estimation in the Gaussian graphical model}.
\newblock \bibinfo{journal}{\emph{Biometrika}} \bibinfo{volume}{94}, \bibinfo{number}{1} (\bibinfo{year}{2007}), \bibinfo{pages}{19--35}.
\newblock


\end{thebibliography}

\appendix

\section{Math Proof}
\subsection{Proof of Theorem~\ref{MDP}}
\label{appendix:MDP}
We start by several important lemmas, which will be useful in the proofs.

\begin{lemma}~\cite{dong2019gaussian}
\label{lem:lemma1}
$\M$ is $(\varepsilon, \delta)-DP$ for all $\varepsilon\ge0$,
    \begin{equation}
    \label{eq:dual}
        \delta(\varepsilon) = \Phi\left(-\frac{\varepsilon}{\mu} + \frac{\mu}{2} \right) - e^{\varepsilon}\Phi\left(-\frac{\varepsilon}{\mu} - \frac{\mu}{2} \right),
    \end{equation}
if and only if $\M$ is $\mu$-GDP, where $\Phi$ is the CDF of normal distribution $\N(0,1)$. Or equivalently, $\mu$-GDP and $(\varepsilon, \delta)-DP$ are primal-dual.  
\end{lemma}

As shown in \cite{dong2019gaussian}, $\mu$-GDP and $(\varepsilon, \delta)-DP$ are actually primal-dual, i.e. $\mu$-GDP can be seen as convex conjugate of collections of $(\varepsilon, \delta)-DP$, and serves as the \textbf{upper bound} of all possible $(\varepsilon, \delta)-DP$ relaxations. Lemma~\ref{lem:lemma1} bridges the two types of differential privacy. In particular, as introduced in Section 3.2, 
$\M$ is $\mu$-GDP implies distinguishing $\mathcal{M}(\dX),\mathcal{M}(\dX^{\prime})$ is more difficult than distinguishing $\mathcal{N}(0,1), \mathcal{N}(\mu,1)$, and is quantified by $G_{\mu} = T(\mathcal{N}(0,1), \mathcal{N}(\mu,1))$. Therefore, we have to show $\M$ is Gaussian differentially private. 

\begin{proposition}
     \label{prop:composition}
    $\mathcal{M}(\dX)$ can be seen as the composition of randomized mechanism $\mathcal{M}_1$ and a post processing mechanism $\text{Proc}$, i.e. $\mathcal{M} = \text{Proc} \circ \mathcal{M}_1$ where
    \begin{align}
        &\mathcal{M}_1(\dX) = f(\dX) + \frac{1}{n}(\dE^{\top}\dX + \dX^{\top}\dE), \\ 
        &\text{Proc}(\boldsymbol{Y}) = \boldsymbol{Y} + \frac{1}{n}( \dE^{\top}\dE).
    \end{align}
\end{proposition}
   
\begin{lemma}~\cite{dong2019gaussian}
\label{lem:composition}
Assume $\mathcal{M}_1$ is $\mu_1$-GDP. Then any post-processing 
$\mathcal{M} = \mathrm{Proc} \circ \mathcal{M}_1$ 
is also $\mu_1$-GDP. Equivalently, post-processed distributions 
can only become more difficult to distinguish than the original distributions.
\end{lemma}

\begin{lemma}
\label{lem:gdp}
$\mathcal{M}_1$ is $\Delta_f(\sigma\sqrt{2C})^{-1}$-GDP, where
\[
\Delta_f = \sup_{\dX,\dX'} 
\| f(\dX) - f(\dX') \|
\]
denotes the global sensitivity over neighboring datasets 
$\dX,\dX'$ that differ in exactly one element, and
\[
C = \min_{k \in \{1,2,\ldots,p\}} 
\frac{1}{n^2}\sum_{l=1}^n \dX_{lk}^2,
\]
i.e., the minimum squared column sum of $\dX$.

\begin{proof}
    We will analyze each randomization by adding to the components of the covariance matrix $f(\dX)$. 

    When $i\neq j$, 
    \begin{equation}
    \begin{aligned}
        \frac{1}{n}(\dE^{\top}\dX + \dX^{\top}\dE)_{ij} &= \frac{1}{n}\sum_{l=1}^n(\dE_{li}\dX_{lj} + \dX_{li}\dE_{lj}) \\
        & \sim \mathcal{N}(0,\frac{\sigma^2}{n^2}\sum_{l=1}^n(\dX^2_{li}+\dX^2_{lj}))
    \end{aligned}
    \end{equation}

    Therefore, for two neighboring data matrices $\dX,\dX^{\prime}$ which differ in exactly one element, we can quantify the difficulty of distinguishing $\mathcal{M}_1(\dX)_{ij}, \mathcal{M}_1(\dX^{\prime})_{ij}$:
    \begin{equation}
    \begin{aligned}
        &T(\mathcal{M}_1(\dX)_{ij}, \mathcal{M}_1(\dX^{\prime})_{ij})\\
        &= T(\mathcal{N}(f(\dX)_{ij},\frac{\sigma^2}{n^2}\sum_{l=1}^n(\dX^2_{li}+\dX^2_{lj})), \mathcal{N}(f(\dX^{\prime})_{ij}, \\
        & \quad \frac{\sigma^2}{n^2}\sum_{l=1}^n(\dX^2_{li}+\dX^2_{lj}))) \\
        &= T(\mathcal{N}(0,\frac{\sigma^2}{n^2}\sum_{l=1}^n(\dX^2_{li}+\dX^2_{lj})), \mathcal{N}(|f(\dX)_{ij}-f(\dX^{\prime})_{ij}|, \\
        & \quad \frac{\sigma^2}{n^2}\sum_{l=1}^n(\dX^2_{li}+\dX^2_{lj}))) \\
        &\ge T(\mathcal{N}(0,\frac{\sigma^2}{n^2}\sum_{l=1}^n(\dX^2_{li}+\dX^2_{lj})), \mathcal{N}(\Delta_f,\frac{\sigma^2}{n^2}\sum_{l=1}^n(\dX^2_{li}+\dX^2_{lj}))) \\
        &= T(\mathcal{N}(0,1), \mathcal{N}(\frac{\Delta_f}{\sigma\sqrt{\frac{1}{n^2}\sum_{l=1}^n(\dX^2_{li}+\dX^2_{lj})}},1)) \\
        &\ge T(\mathcal{N}(0,1), \mathcal{N}(\frac{\Delta_f}{\sigma\sqrt{2C}},1)) = G_{\frac{\Delta_f}{\sigma\sqrt{2C}}},
    \end{aligned}
    \end{equation}
    where the first ``$\ge$'' is because $|f(\dX)_{ij}-f(\dX^{\prime})_{ij}| \le \Delta_f = \sup\limits_{\dX,\dX^{\prime}} | f(\dX) - f(\dX^{\prime}) |$, and the second ``$\ge$'' is because $\frac{1}{n^2}\sum_{l=1}^n(\dX^2_{li}+\dX^2_{lj}) \ge 2C$. 

    Similarly, when $i = j$, 
    \begin{equation}
    \begin{aligned}
        \frac{1}{n}(\dE^{\top}\dX + \dX^{\top}\dE)_{ii} & = \frac{1}{n}\sum_{l=1}^n2\dE_{li}\dX_{lj} \\
        & \sim \mathcal{N}(0,\frac{4\sigma^2}{n^2}\sum_{l=1}^n\dX^2_{li})
    \end{aligned}
    \end{equation}

    In this case, 
    \begin{equation}
    \begin{aligned}
        &T(\mathcal{M}_1(\dX)_{ii}, \mathcal{M}_1(\dX^{\prime})_{ii})\\
        &= T(\mathcal{N}(f(\dX)_{ii},\frac{4\sigma^2}{n^2}\sum_{l=1}^n\dX^2_{li}), \mathcal{N}(f(\dX^{\prime})_{ii},\frac{4\sigma^2}{n^2}\sum_{l=1}^n\dX^2_{li})) \\
        & = T(\mathcal{N}(0,\frac{4\sigma^2}{n^2}\sum_{l=1}^n\dX^2_{li}), \mathcal{N}(|f(\dX)_{ij}-f(\dX^{\prime})_{ij}|, \\
        & \quad \frac{4\sigma^2}{n^2}\sum_{l=1}^n\dX^2_{li})) \\
        &\ge T(\mathcal{N}(0,\frac{4\sigma^2}{n^2}\sum_{l=1}^n\dX^2_{li}), \mathcal{N}(\Delta_f,\frac{4\sigma^2}{n^2}\sum_{l=1}^n\dX^2_{li})) \\
        &= T(\mathcal{N}(0,1), \mathcal{N}(\frac{\Delta_f}{2\sigma\sqrt{\frac{1}{n^2}\sum_{l=1}^n\dX^2_{li}}},1)) \\
        &\ge T(\mathcal{N}(0,1), \mathcal{N}(\frac{\Delta_f}{2\sigma\sqrt{C}},1)) \\
        & \ge T(\mathcal{N}(0,1), \mathcal{N}(\frac{\Delta_f}{\sigma\sqrt{2C}},1)) \\
        &= G_{\frac{\Delta_f}{\sigma\sqrt{2C}}}.
    \end{aligned}
    \end{equation}
\end{proof}
\end{lemma}

Lemma~\ref{lem:lemma1},~\ref{lem:composition},~\ref{lem:gdp} and Proposition~\ref{prop:composition} jointly suggest that $\M$ is $(\varepsilon, \delta)$-DP. Specifically, substituting $\mu$ in (\ref{eq:dual}) by $\frac{\Delta_f}{\sigma\sqrt{2C}}$ yields the results in Theorem~\ref{MDP}.

\subsection{Proof of Theorem~\ref{tho:nontrivial}}
\label{apendix:tho:nontrivial}
\begin{proof}
    Given $\prcs_{ij} = 0$, suppose the shortest path connecting $i,j$ is $i\rightarrow  s_{1}\rightarrow s_{2}\cdots\rightarrow  s_{k}\rightarrow j$. Then, by definition, $\prcs_{is_1}\prcs_{s_1s_2}\cdots \prcs_{s_kj} \neq 0$. Any path shorter than $\{i,s_1,\cdots,s_k,j\}$ cannot connect $i,j$, which indicates:
    \begin{align}
        \prcs^{2}_{ij} &= \sum_{h_1}\prcs_{ih_1}\prcs_{h_1j} = 0 \\
        \cdots\\
        \prcs^{k}_{ij} &= \sum_{h_1}\cdots\sum_{h_{k-1}}\prcs_{ih_1}\prcs_{h_1h_2}\cdots \prcs_{h_{k-1}j} = 0\\
        \prcs^{k+1}_{ij} &= \sum_{h_1}\cdots\sum_{h_k}\prcs_{ih_1}\prcs_{h_1h_2}\cdots \prcs_{h_kj}\neq 0
    \end{align}

On the other hand, if $\sigma^2 < \Vert \prcs \Vert^{-1}$, it follows $\Vert -\sigma^2 \prcs \Vert<1$, then it holds 
\begin{align}
\lim\limits_{n\rightarrow\infty}\left[\I - (\I + \sigma^2\prcs)\right]^n = \lim\limits_{n\rightarrow\infty}\left[-\sigma^2\prcs\right]^n = \zero.
\end{align}

Therefore, through Neumann series expansion, 
    \begin{align}
        \left(\cov+\sigma^2 \I\right)^{-1}_{ij} &= \left(\prcs^{-1}\left(\I + \sigma^2 \prcs\right)\right)_{ij}^{-1} \\
        &=\left(\prcs\cdot\sum_{k=0}^{\infty}\left(-\sigma^2\prcs\right)^k\right)_{ij}\\
        &=\prcs_{ij} - \sigma^2 \prcs^2_{ij} +  \sigma^4 \prcs^3_{ij}-\cdots \\
        &=(-1)^k\sigma^{2k}\prcs^{k+1}_{ij} + \gO(\sigma^{2k})\\
        &\neq 0. 
    \end{align}
    
    Equivalently, $\tilde{\prcs}_{ij}\neq0$ if $\prcs_{ij} = 0$.

    As a toy example, let the precision matrix be
\begin{equation}
    \prcs = \left[
    \begin{array}{ccccc}
     1  &  0.3   &     0   &  0.25 \\
    0.3   &  1    &  -0.1     &     0 \\
         0   & -0.1   &  1    &    0 \\
         0.25  &   0  & 0   & 1   
    \end{array}
    \right].
\end{equation}

After adding Gaussian noise with $\sigma^2 = 0.3$, the perturbed precision matrix become,
\begin{equation}
    \tilde{\prcs} = \left[
    \begin{array}{cccc}
    0.75  &  0.18  &  0.0041  &  0.15 \\
    0.18  &  0.76  & -0.06  & -0.01 \\
    0.0041  & -0.06  &  0.77  & -0.0002 \\
    0.15  & -0.01  & -0.0002  &  0.76
    \end{array}
    \right].
\end{equation}
\end{proof}

\subsection{Proof of Theorem~\ref{tho:pgl}}
\label{apendix:tho:pgl}
\begin{proof}
    We have $ \displaystyle \tilde{\ecov} = \frac{1}{n} (\dX + \dE)^T (\dX + \dE)$, where $\dX \in \mathbb{R}^{n \times p}$ is the raw data and $\dE \in \mathbb{R}^{n \times p}$ is the Gaussian perturbation matrix with i.i.d.\ entries $\dE_{li} \sim \mathcal{N}(0,\sigma^2)$. It then suffices to analyze the trace term,
    \begin{align}
        &\Tr((\tilde{\ecov}-\sigma^2\I)\Theta) \\
        &= \Tr\left(\frac{1}{n} (\dX + \dE)^T (\dX + \dE)\Theta - \sigma^2\I\Theta\right) \\
        &= \Tr((\ecov-\sigma^2\I)\Theta)\\
        &+ \frac{1}{n}\Tr((\dX^T\dE + \dE^T\dX + \dE^T\dE)\Theta)
    \end{align}
    in which 
    \begin{align}
        &\Tr(\dE^T\dE\Theta) \\
        &= \Tr((\dE^T\dE)^T\Theta) \\
        &= \sum^{p}_{i,j=1}\left(\dE^T\dE\right)_{ij}\Theta_{ij} \\
        &= \sum^{p}_{i,j=1}\left(\sum^n_{l=1}\dE_{li}\dE_{lj}\right)\Theta_{ij}.
    \end{align}
    On the other hand,
    \begin{align}
        &\Tr(\dX^T\dE\Theta + \dE^T\dX\Theta) \\
        &= \Tr(\dX^T\dE\Theta) + \Tr(\dE^T\dX\Theta)\\
        &= \Tr(\dX^T\dE\Theta) + \Tr(\dX^T\dE\Theta^T)\\
        &= \Tr(\dX^T\dE(\Theta+\Theta^T)) \\
        & = \Tr((\dE^T\dX)^T(\Theta+\Theta^T)) \\
        &= \sum^{p}_{i,j=1}\left(\dE^T\dX\right)_{ij}\left(\Theta_{ij} + \Theta^T_{ij}\right) \\
        &= \sum^{p}_{i,j=1}\left(\sum^{n}_{l=1}\dE_{li}\dX_{lj}\right)\left(\Theta_{ij} + \Theta^T_{ij}\right).
    \end{align}
    
    Now we analyze the expectation term $\E[\dE_{li}\dE_{lj}]$ and $\E[\dE_{li}\dX_{lj}]$. It is evident $\forall i,j, \E[\dE_{li}\dX_{lj}] = 0$. 
    We can see $\E[\dE_{li}\dE_{lj}] = 0$ when $i\neq j$. In the case $i=j$, $\dE_{li}\dE_{lj}$ is Chi-square distribution with $1$ degree of freedom scaled by $\sigma^2$, and
    $\displaystyle \E[\dE_{li}\dE_{lj}] = \sigma^2$. Collectively,
    \begin{align}
        &\E[\frac{1}{n}\Tr(\dE^T\dE\Theta + \dX^T\dE\Theta + \dE^T\dX\Theta)] \\
        &= \frac{1}{n}\sum^{p}_{i=j} n\sigma^2 \Theta_{ij} = \sum^{p}_{i=j}\sigma^2 \Theta_{ij} \\
        &= \Tr(\sigma^2\I\Theta).
    \end{align}

    Consequently, $\displaystyle\E\left[\Tr((\tilde{\ecov}-\sigma^2\I)\Theta)\right] = \Tr(\ecov\Theta)$. It follows immediately $\displaystyle J(\Theta; \ecov) = \E\left[J(\Theta;\tilde{\ecov}-\sigma^2\I)\right]$.
\end{proof}

\subsection{Proof of Theorem~\ref{tho:DMDP}}
\label{apendix:tho:DMDP}
We start by introducing some important facts and lemma about $\dN(\mu,\sigma^2)$, which will be useful.

\begin{proposition}[Discretization on $t\Z$~\cite{canonne2020discrete}]
\label{prop:dis}
If the Discretization of Gaussian is on $t\Z:=\{tz:z\in\Z\}$ instead of $\Z$, this distribution can be denoted as $\dNt(0,\sigma^2)$.  If $\nX\sim\dN(0,\sigma^2)$, then $t\nX\sim\dNt(0,t^2\sigma^2)$. More importantly, as pointed out in~\cite{peikert2016decade}, the sum of independent discrete Gaussians (over the same support) is still a discrete Gaussian.
\end{proposition}

\begin{proposition}[Tail bound for discrete Gaussian~\cite{canonne2020discrete}]
\label{prop}
For $\sigma > \frac{t}{\sqrt{2\pi}}$ and any $x\in\Z$, $t>0$,
\begin{equation}
   \Prob[\nY \le x] > \Prob[\nX \le x] \ge \Prob[\nY \le x-t],
\end{equation}
where $\nX\sim \dNt(0, \sigma^2), \nY\sim \N(0, \sigma^2)$.
\end{proposition}

By the above proposition, we can relax the trade-off function between discrete Gaussian and continuous Gaussian, and get the following lemma.

\begin{lemma}
\label{lem:dgdp}
The difficulty to distinguish $\dNt(0, \sigma^2), \dNt(\Delta, \sigma^2)$ is at least harder than $G_{\frac{\Delta + 2}{\sigma}}$. 
Equivalently,
\[
T(\dNt(0, \sigma^2), \dNt(\Delta, \sigma^2)) \ge G_{\frac{\Delta + 2t}{\sigma}}
\].

\begin{proof}
By definition of the trade-off function,
for input significance level $\alpha$, it measures the level of type II error:
\begin{align}
    &T(\dNt(0, \sigma^2), \dNt(\Delta, \sigma^2))(\alpha)\\
    &= \Prob[\nX \le x_{\alpha}], \nX\sim \dNt(\Delta, \sigma^2) \\
    &= \Prob[\nX \le x_{\alpha} - \Delta], \nX\sim \dNt(0, \sigma^2) \\
    &\ge \Prob[\nY \le x_{\alpha} - \Delta - t], \nY\sim \N(0, \sigma^2),
\end{align}
where the inequality is because Proposition~\ref{prop}, $x_{\alpha}$ such that $\Prob[\nX > x_{\alpha}]=\alpha, \nX\sim \dNt(0, \sigma^2)$.
On the other hand, 
\begin{align}
    &T(\N(0, \sigma^2), \N(\Delta+2t, \sigma^2))(\alpha)\\
    &= \Prob[\nY \le y_{\alpha}], \nY\sim \N(\Delta+2t, \sigma^2) \\
    &= \Prob[\nY \le y_{\alpha} - \Delta - 2t], \nY\sim \N(0, \sigma^2),
\end{align}
where $y_{\alpha}$ such that $\Prob[\nY > y_{\alpha}]=\alpha, \nY\sim \N(0, \sigma^2)$.
By Proposition~\ref{prop}, $\alpha = \Prob[\nX > x_{\alpha}]= \Prob[\nY > y_{\alpha}], \nX\sim \dNt(0, \sigma^2), \nY\sim \N(0, \sigma^2)$  leads to $|x_{\alpha} - y_{\alpha}|<t$, we have $x_{\alpha} - \Delta - t > y_{\alpha} - \Delta - 2t$, then for $\nY\sim \N(0, \sigma^2)$,
\begin{equation}
    \Prob[\nY \le x_{\alpha} - \Delta - t] \ge \Prob[\nY \le y_{\alpha} - \Delta - 2t].
\end{equation}

Together,
\begin{align}
    &T(\dNt(0, \sigma^2), \dNt(\Delta, \sigma^2))(\alpha)\\
    &\ge T(\N(0, \sigma^2), \N(\Delta+2t, \sigma^2))\\
    &= T(\N(0, 1), \N((\Delta + 2t)/\sigma, 1)) \\
    &= G_{\frac{\Delta+2t}{\sigma}}.
\end{align}

\end{proof}
\end{lemma}

\begin{proposition}
\label{prop:decompose}
Assume $\x_{li}\in\dX\neq 0$, and  Let $K = \frac{1}{n}\min_{\x_{li}\in\dX}|\x_{li}|$. The mechanism $\mathcal{M}$ can be represented as composition of mechanism $\mathcal{M}_1$ and a post processing mechanism $\text{Proc}$, i.e. $\mathcal{M} = \text{Proc} \circ \mathcal{M}_1$,
    \begin{equation}
    \begin{aligned}
        &\mathcal{M}_1(\dX) = f(\dX) + (K\boldsymbol{I}_{p\times n} \dEz + K\dEz^{\top}\boldsymbol{I}_{n\times p}) , \\ 
        &\text{Proc}(\boldsymbol{Y}) = \boldsymbol{Y} + \frac{1}{n}( \dEz^{\top}\dEz) + \frac{1}{n}(\dEz^{\top}\dX + \dX^{\top}\dEz)\\
        & -(K\boldsymbol{I}_{p\times n} \dEz + K\dEz^{\top}\boldsymbol{I}_{n\times p}).
    \end{aligned}
    \end{equation}
\end{proposition}

\begin{lemma}
\label{lem:d}
In the discrete Gaussian case, $\mathcal{M}_1$ is
$\frac{\Delta_f + 2K}{\sqrt{2n}\,K\sigma}$-GDP, where
\[
\Delta_f = \sup_{\dX,\dX'} \| f(\dX) - f(\dX') \|
\]
denotes the global sensitivity, where the supremum is taken over all neighboring datasets $X, X'$ that differ in exactly one element,
\begin{proof}
We analyze each randomization by adding noise to the components of the covariance matrix $f(\dX)$.

When $i\neq j$,
\begin{align}
(K\boldsymbol{I}_{p\times n}\dEz + K\dEz^{\top}\boldsymbol{I}_{n\times p})_{ij}
&= K\sum_{l=1}^{n}\bigl((\dEz)_{li} + (\dEz)_{lj}\bigr) \\
&\sim \mathcal{N}_{K\mathbb{Z}}\!\left(0,\,2nK^2\sigma^2\right).
\end{align}
Then, by Lemma~\ref{lem:dgdp}, this mechanism is
$\frac{\Delta_f + 2K}{\sqrt{2n}\,K\sigma}$-GDP.

When $i=j$,
\begin{align}
(K\boldsymbol{I}_{p\times n}\dEz + K\dEz^{\top}\boldsymbol{I}_{n\times p})_{ii}
&= 2K\sum_{l=1}^{n}(\dEz)_{li} \\
&\sim \mathcal{N}_{K\mathbb{Z}}\!\left(0,\,4nK^2\sigma^2\right).
\end{align}
Then, by Lemma~\ref{lem:dgdp}, this mechanism is 
$\frac{\Delta_f + 2K}{2\sqrt{n}K\sigma}$-GDP,
which is stronger than 
$\frac{\Delta_f + 2K}{\sqrt{2n}K\sigma}$-GDP.
Therefore, taking the worst-case (off-diagonal) bound yields the overall GDP guarantee.

\end{proof}
\end{lemma}

Lemma~\ref{lem:lemma1},~\ref{lem:composition},~\ref{lem:dgdp},~\ref{lem:d}, Proposition~\ref{prop:decompose} jointly give the theorem about $\M$'s utility in privacy shown in Theorem \ref{tho:nontrivial}.

\end{document}